\pdfoutput=1
\documentclass[11pt]{article}
\usepackage{indentfirst}
\usepackage{ACL2023}

\usepackage{times}
\usepackage{latexsym}

\usepackage[T1]{fontenc}
\usepackage{algorithm}
\usepackage{algpseudocode}
\usepackage{amsmath}

\usepackage[utf8]{inputenc}

\usepackage{microtype}

\usepackage{inconsolata}
\usepackage[utf8]{inputenc}
\usepackage{graphicx}

\usepackage{makecell}
\usepackage{subcaption}

\title{Ex3: Automatic Novel Writing by Extracting, Excelsior and Expanding}

\author{%
    Lei~Huang\textsuperscript{1,2,3~\textdagger} \quad
    Jiaming~Guo\textsuperscript{1~\textdagger}\thanks{Corresponding Author.} \quad 
    Guanhua~He\textsuperscript{1,2,3~\textdagger} \quad
    Xishan~Zhang\textsuperscript{1,2} \quad
    Rui~Zhang\textsuperscript{1}\quad \\
    \textbf{Shaohui~Peng\textsuperscript{4}\quad 
    Shaoli~Liu\textsuperscript{2}\quad
    Tianshi~Chen\textsuperscript{2}} \\ 
    \textsuperscript{1} SKL of Processors, Institute of Computing Technology, CAS, Beijing, China \\
    \textsuperscript{2} Cambricon Technologies\\
    3 University of Chinese Academy of Sciences, UCAS, Beijing, China \\
    \textsuperscript{4} Intelligent Software Research Center, Institute of Software, CAS, Beijing, China \\
    {\tt\small  \{huanglei22s,guojiaming\}@ict.ac.cn,heguanhua22@mails.ucas.ac.cn,\{zhangxishan,zhangrui\}@ict.ac.cn} \\
    {\tt \small  \{liushaoli,tchen\}@cambricon.com }  \\
}

\begin{document}
\maketitle
\def\thefootnote{\textdagger}\footnotetext{These authors contributed equally to this work.}\def\thefootnote{\arabic{footnote}}
\begin{abstract}
Generating long-term texts such as novels using artificial intelligence has always been a challenge. A common approach is to use large language models (LLMs) to construct a hierarchical framework that first plans and then writes. 
Despite the fact that the generated novels reach a sufficient length, they exhibit poor logical coherence and appeal in their plots and deficiencies in character and event depiction, ultimately compromising the overall narrative quality.
In this paper, we propose a method named Extracting Excelsior and Expanding. 
Ex3 initially extract structure information from raw novel data. By combining this structure information with the novel data, an instruction-following dataset is meticulously crafted. This dataset is then utilized to fine-tune the LLM, aiming for excelsior generation performance. In the final stage, a tree-like expansion method is deployed to facilitate the generation of arbitrarily long novels.
Evaluation against previous methods showcases Ex3's ability to produce higher-quality long-form novels.

\end{abstract}
\section{Introduction}
% With the high academic interest and application value of automatic long-form text generation, a series of works propose to generate long-form text with artificial intelligence. 

The realm of automatic long-form novel generation through artificial intelligence holds significant appeal both in academic and practical domains, yet has posed persistent challenges for a long time. %This demands not only a comprehensive grasp of linguistic and commonsense knowledge but also the adeptness to plan for the coherent construction of extensive textual content. 
This demands not only solid language expression ability \citep{DBLP:journals/nle/Stede96,yang2022re3} but also strong logical thinking and organizational skills to conceive an appealing and coherent storyline.
%In recent times, large language models (LLMs) have showcased strong capabilities across diverse text generation tasks. Consequently, leveraging LLMs for the purpose of long-form novel generation emerges as a promising and viable solution.

In recent times, large language models (LLMs) possess excellent language expression ability and basic common sense reasoning skills, thus leveraging LLMs for narrative generation is promising. However, generating long novels is still challenging, because of the inherent fixed-size context design in the Transformer architecture.
%Owing to the inherent fixed-size context design in the Transformer architecture, generating a long novel in one shot with LLMs becomes difficult. 
%For human writers, a common practice involves the creation of a plan or outline prior to commencing a novel. Inspired by this, 
%Some works propose the hierarchical generation approach.use A general-purpose LLMs and first prompt the LLM to generate outlines, followed by subsequent prompts to generate novels in accordance with the previously devised plans.
Some works~\cite{yang2022re3,yang2022doc,zhou2023recurrentgpt} propose the hierarchical generation approach, where general-purpose LLMs are used to directly generate outlines, and followed by subsequent prompts to generate novels in accordance with the previously devised outlines. 
% Despite being able to produce novels that are sufficiently long, these methods have issues with the logical coherence of the storyline and the inconsistency of character development, resulting in novels that appear highly unnatural.
Despite being able to produce novels that are sufficiently long, these methods have issues with the logical coherence and appeal of the storyline and the deficiencies in character and event depiction, resulting in novels that appear highly unnatural.  The reason is that the general-purpose LLM has insufficient planning and linguistic capabilities in the field of novel writing, which cannot be solved simply by prompt engineering. We believe that designing AI writing systems should not only consider how people write but also how people learn to write.

%However, the lack of planning ability and novel-style linguistic ability of the general-purpose LLMs leads to the limited quality of the resulting novel. The general-purpose LLMs have not been trained to learn the structural information of the novel such as the outlines and the change of character settings, which is crucial for the coherence and appeal of the novel.  Furthermore, the writing style generated by these general-purpose LLMs often fails to align with the style demanded by a novel.
How do human writers learn to construct reasonable storylines and improve their linguistic capacity? Reading is one of the best ways to become a better writer. The more you read and summarize, the better you will write. Inspired by this, we propose to learn organization skills and linguistic capacity from human writing. 
We first organize outlines from existing novels and then conduct the reverse process, that is, creating an outline and developing the outline into a novel. It's like a regular rewriting exercise that writers do.

%To address these limitations, 
In particular, we propose the Extract Excelsior and Expand framework (Ex3) to automatically construct storyline by learning from raw novels, and generate new novels that imitate the linguistic style of a specific genre.
%In particular, we propose the Extract Excelsior and Expand framework (Ex3) to automatically learn planning and linguistic capabilities from raw novels and generate novels with higher quality. 
%Ex3 first \textbf{Extracts} structural information from the raw novel data. It is expensive to obtain the structural information in the novel manually, so we turn to use self-instructing method. In order to obtain different levels of structural information, we use similarity to divide the text and design an iterative summarising method. In addition, we also used the LLM for entity extraction and corresponding it to the outline of different levels. 
Ex3 first \textbf{Extracts} structural information from the raw novel data using self-instructing method. In order to obtain different levels of structure information, we divide the text into sections according to content similarity. Additionally, we utilize entity extraction techniques to develop rich profiles of the characters, further enhancing the narrative's depth and coherence.
In the \textbf{Excelsior} stage, we construct an instruction-following dataset with the extracted structure information and the raw novel. Then we enhance the LLM by finetuning it with the dataset. Through this process, the language used in the generated novels more closely aligns with the stylistic choices made by human novelists, resulting in a more natural and engaging reading experience.
At last,  a tree-like method is used to \textbf{Expand} a premise to a novel.  We use a depth-first approach to generate the novel, combined with entity extraction to make the content of the novel more coherent. Finally, all the leaves of the tree make up the novel. The proposed Ex3 is a self-improvement framework. With the insight that summarizing is easier than expanding, we use the same LLM in three stages, that is, the LLM summarizes the novel to create the dataset and improve itself, which is similar to how human writers learn.

To evaluate the proposed Ex3 for long-term novel generation, we compare its generated novels to similar-length novels generated by several previous methods. We evaluate the novel by both human study and automation metrics. The results demonstrate that Ex3 is able to generate long-term novels with higher quality.

\begin{figure*}[!t]
    \centering
    
    \includegraphics[width=0.87\linewidth]{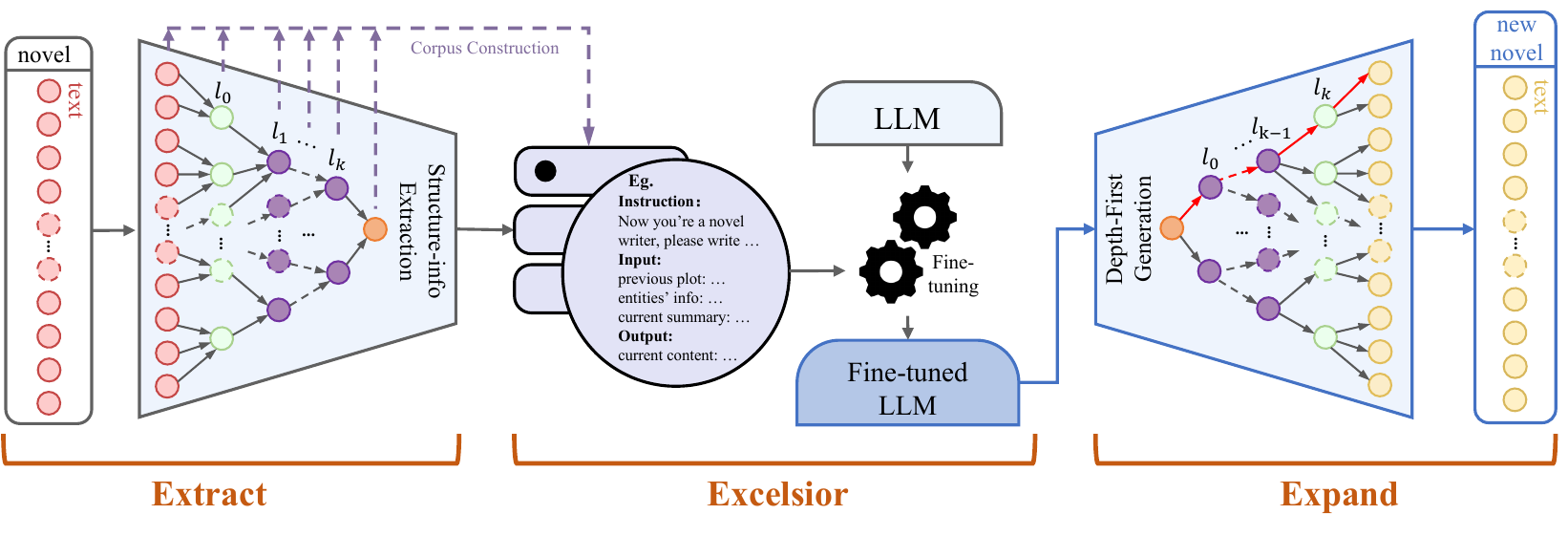}
    %\vspace{-13pt}
    \caption{\textbf{The Ex3 framework for novel writing}. \textit{Extract} for structure information extraction; \textit{Excelsior} for corpus construction and LLM fine-tuning; \textit{Expand} for automatic novel generation.}
    \label{Framework}
   % \vspace{-12pt}
\end{figure*}
\section{Related Work}

\paragraph{Long Context Transformers.}
Large language models face limitations when it comes to writing novels due to their inadequate capacity to effectively manage lengthy texts and retain extensive contextual memory and associations \citep{qiu2019blockwise, zhou2023recurrentgpt}. 
% Proposed solutions in related research can be broadly categorized into three main approaches: sparse attention mechanisms, position encoding optimizations, and the introduction of caching mechanisms. Many researchers have observed that the computational resources and memory consumption of self-attention mechanisms grow quadratically with the input sequence length \citep{beltagy2020longformer}. 
Several works make efforts to design efficient attention mechanisms that can increase the models' contextual window size \citep{qiu2019blockwise, kitaev2020reformer, choromanski2020rethinking}. \citet{child2019generating} proposes sparse attention weight matrices, transforming the fully connected self-attention mechanism into a form of local attention to reduce resource consumption. Subsequent works have continuously optimized long sequence modeling, enabling attention mechanisms to achieve linear complexity \citep{zaheer2020big, wang2020linformer}. Some other methods focus on enhancing the encoding of positional information in models \citep{liu2020learning}, with ALiBi \citep{press2021train} as a representative which adds linear biases to the attention mechanism. Some works introduce a caching mechanism to enhance the model's memory capacity to process text information from different context windows \citep{dai2018transformer, bertsch2023unlimiformer}, such as simulating LSTM \citep{wang2019r, bulatov2022recurrent, zhou2023recurrentgpt}. Although these methods can support longer inputs, they do not significantly improve the output length.

\paragraph{Automatic Novel Generation.}

Despite researchers' significant efforts to expand the window size of general-purpose language models (LLMs), the length of text generated in a single output remains insufficient for the requirements of a full-length novel. Consequently, researchers have explored a hierarchical approach \citep{DBLP:conf/acl/LewisDF18, yao2019plan, sun2020summarize, DBLP:conf/naacl/TanYAXH21}, where LLMs are used to directly produce outlines, which are then followed by additional prompts to generate novels based on the previously created outlines. On the one hand, to enhance the coherence of the generated novel's plot in the hierarchical approach, a coarse-to-fine strategy has been proposed \citep{xu2018skeleton, DBLP:conf/acl/FanLD19}. It involves extracting important events or entities and arranging the logical progression of the plot to guide the generation of the novel's story \citep{zhou2018neural, yao2019plan}. On the other hand, to increase the length of the generated novel, a recurrent generation method has been introduced \citep{yang2022re3, yang2022doc, zhou2023recurrentgpt}. This method generates partial content in each iteration using a "plan and write" approach, acquiring multiple rounds to obtain a full-length novel. It incorporates modification mechanisms or memory mechanisms to improve the coherence of the generated text across iterations. However, as the length of the text increases, issues such as inconsistencies and logical confusion still arise.

Existing hierarchical writing methods heavily rely on the linguistic capabilities of large language models themselves, neglecting deeper exploration and learning from the data of novel texts. As a result, it is challenging to simultaneously achieve both the length of the generated text and the coherence of its contextual logic. To enhance Ex3's performance in generating full-length novels, we focus on effectively learning the plot structure and language style of novels from raw data. 

\section{Extract, Excelsior, and Expand}
% The process of writing a novel can be roughly abstracted as a hierarchical extension of tree-like logic, where structure information is implicitly embedded in the development of the plot. 
% Our goal is to enable the LLMs to learn the hierarchical structure information and write novels like humans. Therefore, we have designed three \textit{Ex} processes as shown in Figure \ref{Framework}: \textit{Extract} for extracting structure information from raw novel texts, \textit{Excelsior} for constructing training corpus to fine-tune the LLMs, and \textit{Expand} for generating novels in a human-like writing style automatically. In this section, we will provide a detailed introduction to these three processes.

This section introduces the proposed Extracting Excelsior and Expanding (Ex3) framework. As Figure \ref{Framework} shows, the framework consists of three \textit{Ex} processes: \textit{Extracting} for extracting structure information from raw novel texts, \textit{Excelsior} for constructing training corpus to fine-tune the LLMs, and \textit{Expanding} for generating novels in a human-like writing style automatically.

\subsection{Extracting}
Based on the raw text of the novel, we can reverse-engineer the hierarchical structure information, so as to reconstruct the logical layout and narrative design of the novel's storylines. 
We use a \textit{Grouping and Summarizing} approach to extract hierarchical structure information from bottom to top. In this approach, \textit{Grouping} refers to aggregating consecutive texts with high relevance into a group, as we believe these texts are more likely to come from the same outline. \textit{Summarizing} refers to extracting a higher-level outline of each text group. In this section, we'll introduce the text grouping method based on semantic similarity, as well as the processes of chapter summarizing and constructing structure information for the entire novel text. Moreover, to ensure the logical coherence of key contextual information, we propose the entity information extraction method.

\paragraph{Group the text by Similarity.} 
The window length of a large language model is limited, and it cannot process an entire novel as input at once. Therefore, we need to split the novel text into smaller parts. However, using a fixed-length window for splitting would significantly disrupt the integrity of the text information and semantic logic. To address this issue, we propose a text grouping method based on semantic similarity.

In the process of writing a novel, the stories or scenes unfold in an orderly manner, similar to the progression of shots in a movie. We have noticed that there is a higher semantic relevance between consecutive paragraphs belonging to the same "shot". When there is a transition in the story or scene, the semantic similarity between paragraphs decreases to a lower value. Based on this characteristic, we use text semantic similarity to measure the correlation between paragraphs and group the text according to the story or scene, without compromising the coherence of the semantic information. 
% Specifically, We embed the paragraphs with a pre-trained CoSENT model \citep{text2vec} and then use cosine distance as the similarity.

\begin{figure}[!h]
    \centering
    \includegraphics[scale=0.42]{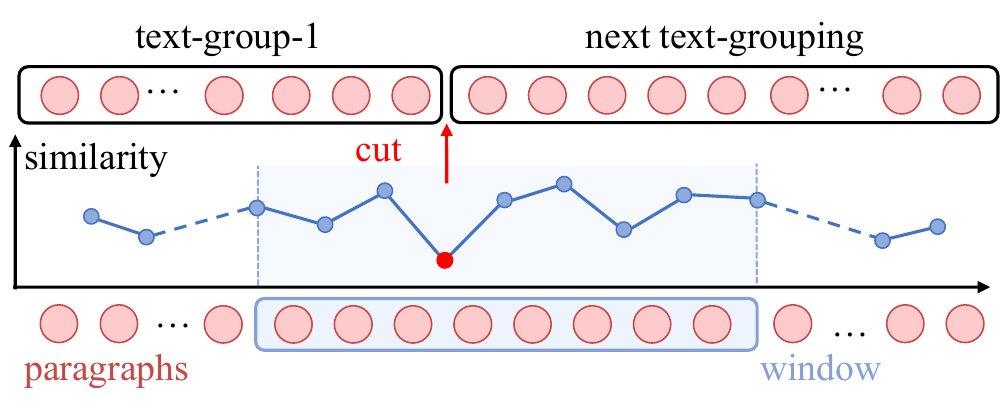}
    % \vspace{-12pt}
    \caption{\textbf{Group the Text by Similarity}. We calculate the semantic similarity scores and segment a window. Then we choose the paragraph corresponding to the minimum score in the window as the partition.}
    \label{GrouptheTextbySimilarity}
   % \vspace{-8pt}
\end{figure}

The process is shown in Figure \ref{GrouptheTextbySimilarity}. Firstly, we calculate the semantic similarity between all adjacent paragraphs in the text. Then, a paragraph window is created based on a length range. Within each window, we select the two paragraphs with the lowest similarity as the partition points. Next, we apply the same operation to the remaining text until the entire article is divided into multiple groups of paragraphs. This method takes paragraphs as the smallest unit and groups the stories or scenes in the text while maintaining semantic integrity. The length range controls the granularity of information in each group, making the partitioning process more flexible.

Specifically, We embed the paragraphs with a pre-trained CoSENT model \citep{text2vec} and then use cosine distance as the similarity. The range of window length is determined by the maximum text length the model can handle. For the Baichuan2-13B-Chat model \citep{baichuan2023baichuan2} that we use in the experiment, whose window length is limited to 2048, we specify a minimum text length (Chinese characters) of 400 and a maximum of 1800 in our experiments.

\paragraph{Chapter Summarizing.}
% Chapters are the core components of the narrative structure and plot development in novel writing and embody the main substance of the text. They not only delineate the unfolding of the storyline but also reflect the literary proficiency of the author. Consequently, chapters are treated as the fundamental units for the structure information extraction. 

Chapters are the core components of the narrative structure and plot development in novel writing and embody the main substance of the text. Consequently, they are treated as the fundamental units for the structure information extraction. 

\begin{figure}[!h]
    \centering
    \includegraphics[scale=0.65]{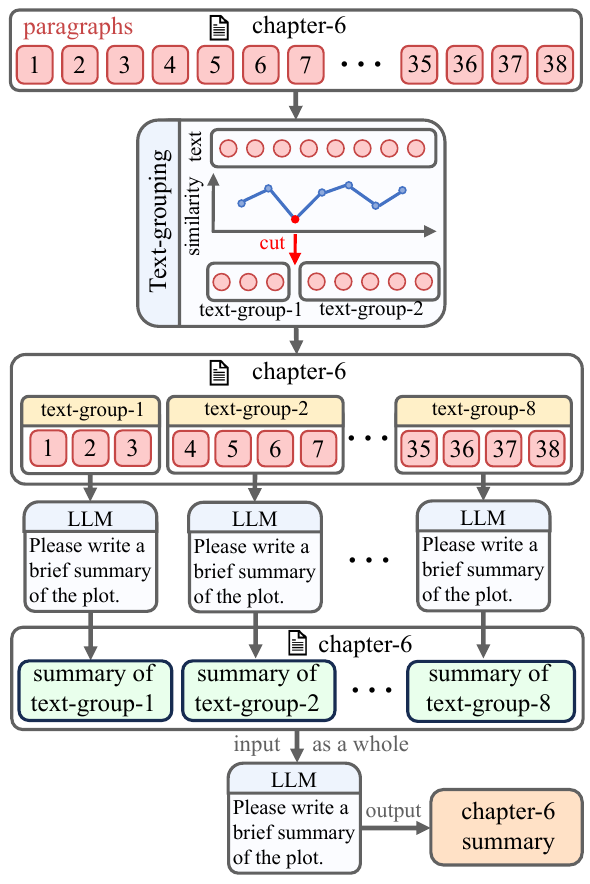}
    %\vspace{-8pt}
    \caption{A brief example of Chapter Summarizing. A chapter consisting of 38 paragraphs is divided into 8 groups, and each group is summarized by LLMs individually. These group summaries are integrated to produce the comprehensive summary of the chapter.}
    %\vspace{-12pt}
    \label{ChapterSummarizing}
\end{figure}

Each chapter typically encompasses a constellation of interrelated events, scenes, or descriptions. 
We first group the paragraphs by semantic similarity within a chapter. Then, we utilize a large language model to generate summaries for each paragraph group. These summaries are then sequentially merged to form a composite overview, which is fed back into the large language model to generate a coherent chapter summary.

As illustrated in Figure \ref{ChapterSummarizing}, this method employs a two-level hierarchical structure for each chapter, which segments the main body of the chapter into smaller sections while adhering to the text window size limits of the LLMs.

\paragraph{Recursively Summarising.}
An outline typically guides the logical progression of a story, and in the case of a long novel, it often consists of multiple hierarchical levels of outlines. These outlines form an expanded structure of a tree-like logic, representing the conceptualization process during novel writing. Starting with chapter summaries, our methodology enables the reverse extraction of outline information across these various hierarchical levels.

% \begin{figure}[!h]
%     \centering
%     \includegraphics[scale=0.55]{Pics/RecursivelySummarizing-shorter.pdf}
%     \vspace{-12pt}
%     \caption{An illustration of Structure-Info Extraction. After applying the method described above to obtain summaries for each chapter in a novel, we group all the chapter summaries by semantic similarity and then generate a summary for each group. We repeat this process until we eventually condense it into a single summary.}
%     \label{RecursivelySummarizing}
%     \vspace{-12pt}
% \end{figure}

\begin{figure}[!h]
    \centering
    \includegraphics[scale=0.55]{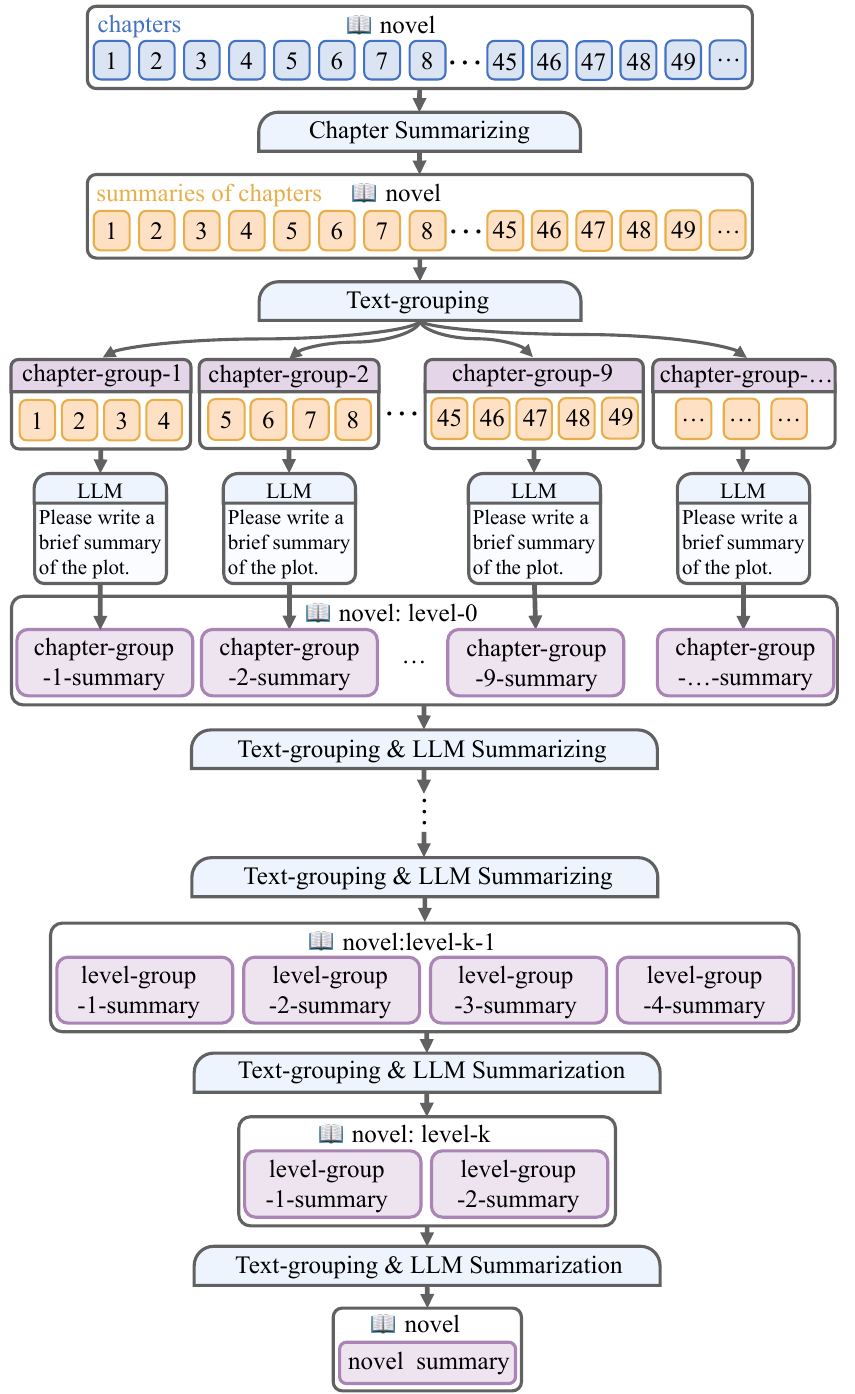}
    %\vspace{-12pt}
    \caption{An illustration of Structure-Info Extraction. After obtaining chapter summaries for the entire novel using the Chapter Summarizing method, we group all the chapter summaries by semantic similarity and then generate a summary for each group. We repeat this process until we eventually condense it into a single summary.}
    \label{RecursivelySummarizing}
    %\vspace{-12pt}
\end{figure}

We utilize a recursive summarizing method to obtain hierarchical structure information (Figure \ref{RecursivelySummarizing}).
Firstly, we utilize the Chapter Summarizing method to extract summaries of each chapter in the novel, which are then integrated in a sequential manner. Next, we employ semantic similarity to segment these summaries into groups, with each containing stories of strong internal coherence. We then use a large language model to summarize each group, generating outlines for the next level. This process can be repeated iteratively to gradually obtain outlines encompassing varying levels of information granularity until the final comprehensive summary of the novel is achieved.
% The termination condition can also be flexibly determined based on the text length or the number of groups.

A termination criterion is required to converge to a final novel summary, which can be flexibly determined based on the previous hierarchy. For example, when there are only two level group summaries remaining, or when the total length of the summaries in the previous hierarchy falls below a threshold set beforehand, the algorithm terminates after collapsing them into a single summary. 
Another option is to consider the semantic similarities between the level group summaries in the previous hierarchy, which can also be constrained using a threshold. If the semantic similarities among the level group summaries are sufficiently high, the algorithm proceeds to generate the final summary. In our experiments, we use the text length as the convergence criterion.
\begin{figure}[!t]
    \centering
    \includegraphics[scale=0.45]{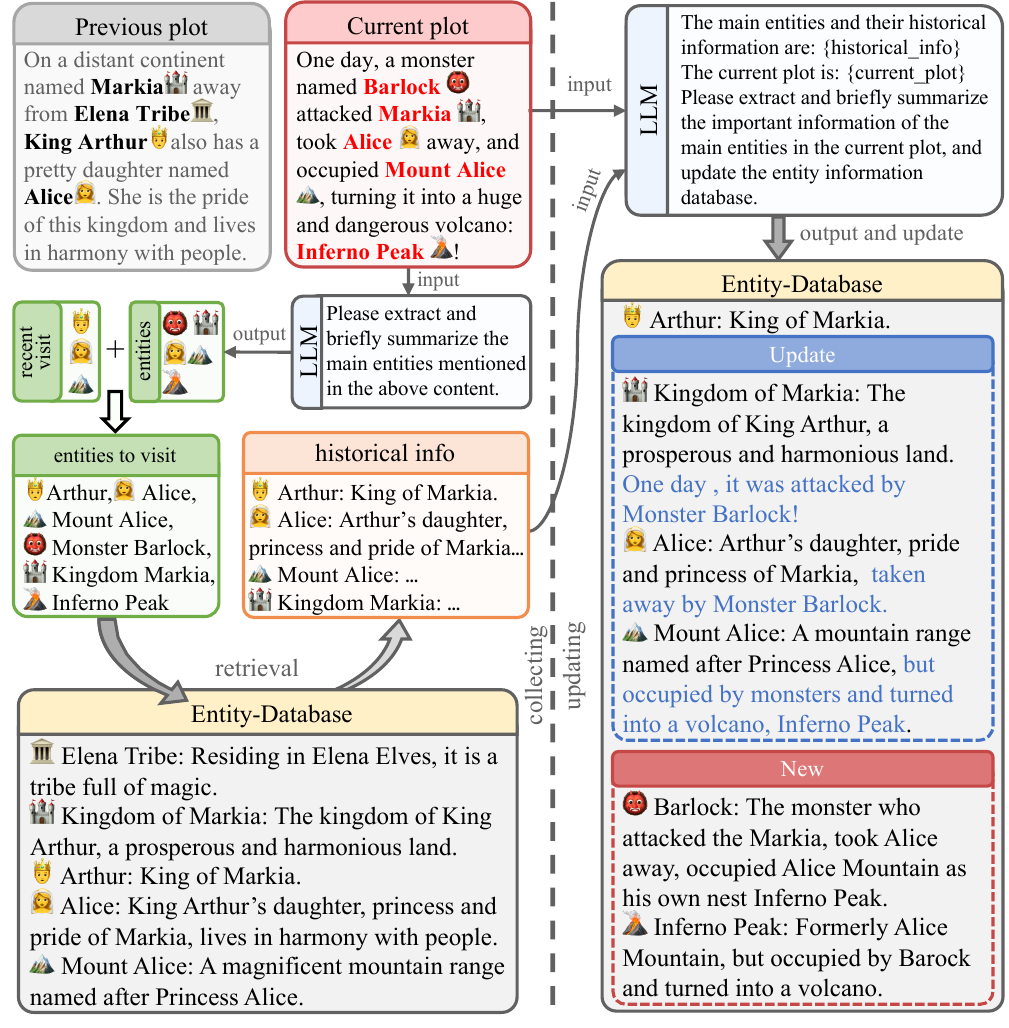}
    %\vspace{-15pt}
    \caption{A simple example of Entity Information Extraction method. When it comes to the current plot, the entities captured by LLMs, along with recent visits, are retrieved from the database to obtain the historical information of relevant entities prior to the current plot events. Then the current plot and historical information are both input to LLMs to obtain the latest entity information, thereby adding new entries or updating the information of related existing entries in the database.}
    \label{EntityExtraction}
    %\vspace{-12pt}
\end{figure}
\paragraph{Entity Extraction}
In order to ensure the alignment of entity information throughout the full text, alleviate the problem of model forgetting, and enhance the coherence between preceding and subsequent contexts, we incorporate the entity information extraction method. Entity extraction refers to the process of identifying entities with specific significance in the text and extracting crucial information associated with them, such as characters, locations, or organizations.

The method maintains an entity information database, focusing on the text of segmented paragraph groups as the primary objects. It tracks and records key information about entities as the novel's narrative unfolds, which is divided into two stages: collection and updating (Figure \ref{EntityExtraction}).
In the collecting stage, we utilize a large language model to identify the entity names in the current text. Together with recently visited entities, we retrieve entries from the entity information database, to construct the historical knowledge of the entities related to the current text.
% In the updating stage, the model leverages the historical knowledge as the background to extract key information about the entities from the current text, thereby updating the entity information database.
In the updating stage, the model leverages historical knowledge as the background to extract key information about the entities from the current text. It does not retain all historical details, but instead selectively discards some past and less important knowledge and incorporates new information, thereby evolving the entity information database with the progress of the story.

\begin{figure}[t]
    \centering
    \includegraphics[scale=0.45]{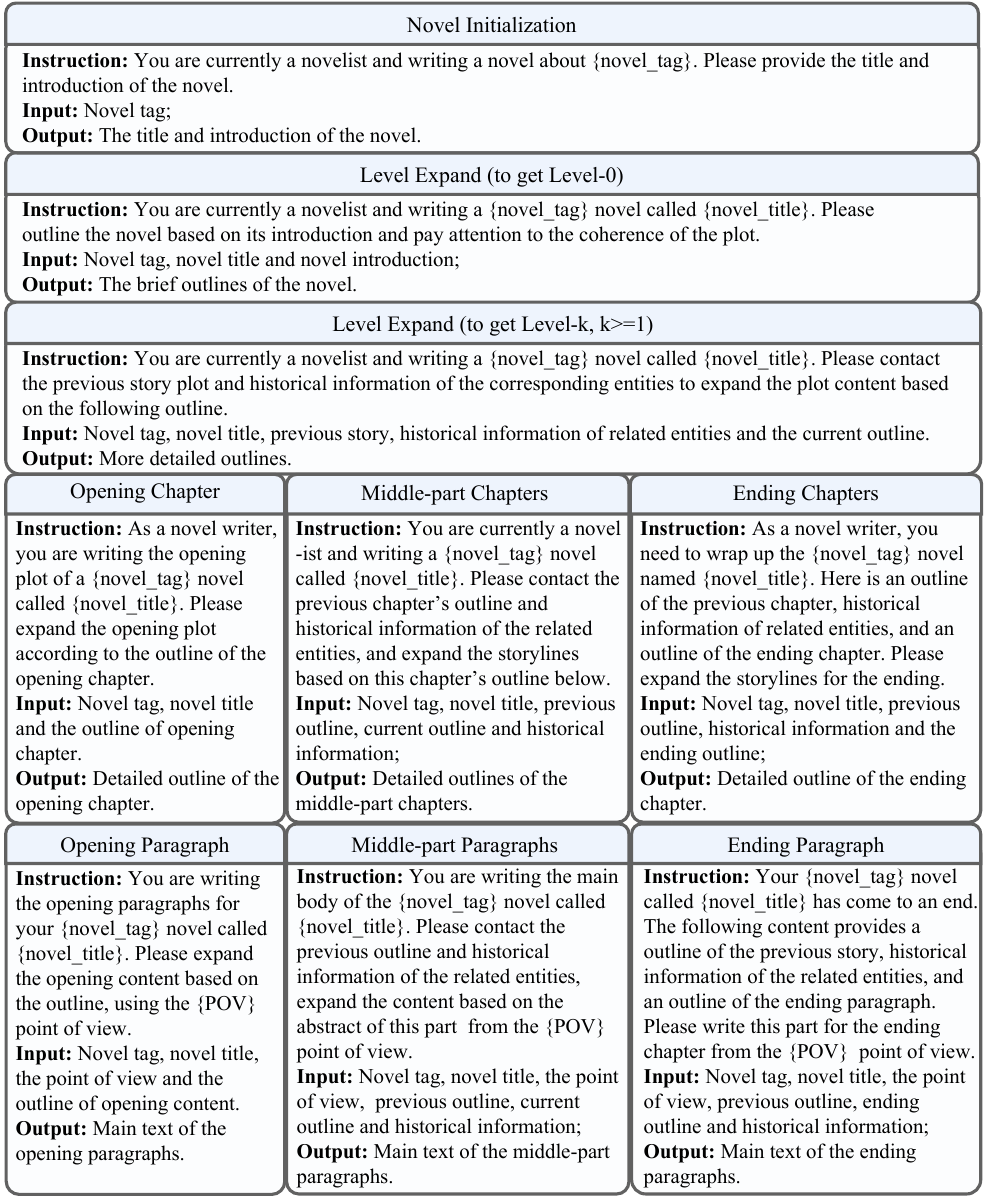}
    %\vspace{-12pt}
    \caption{Five categories of prompts to distinguish different hierarchical levels.}
    \label{Instructions}
   % \vspace{-12pt}
\end{figure}
\subsection{Excelsior}
\begin{figure*}[t]
    \centering
    \includegraphics[width=0.87\linewidth]{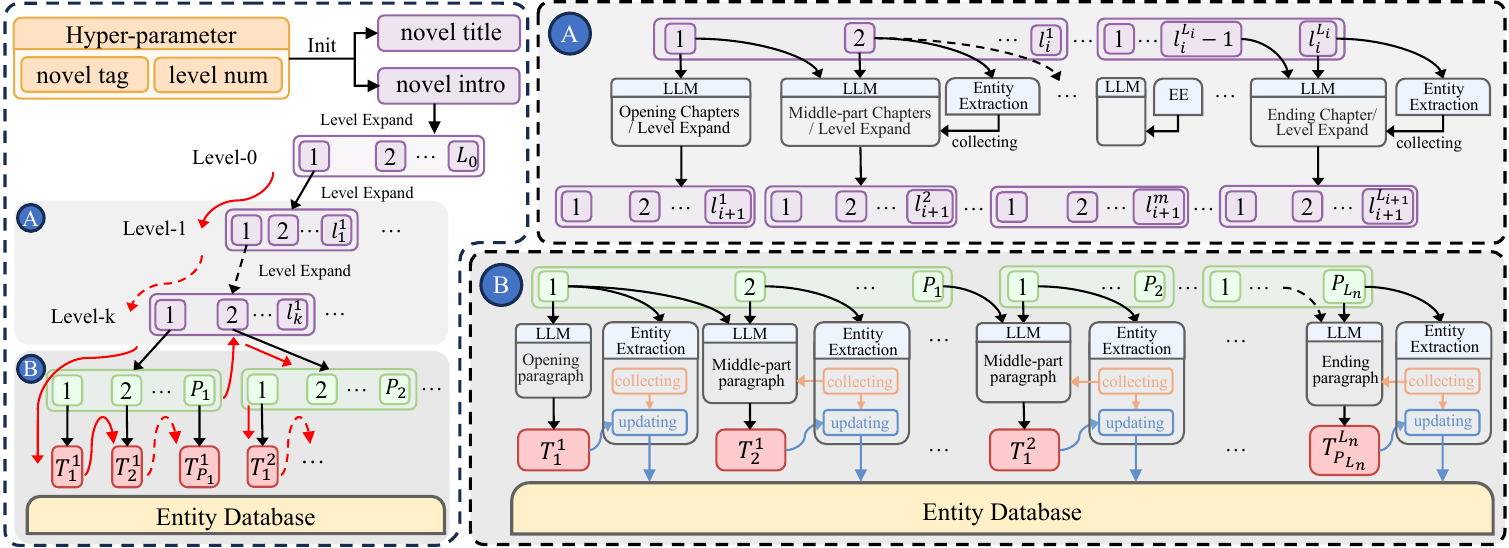}
    %\vspace{-5pt}
    \caption{\textbf{Depth First Generation}. A brief overview of the entire process is depicted in the diagram on the left, illustrating its characteristic of depth-first generation. Subgraph A depicts the expansion of intermediate-level outlines and subgraph B illustrates the expansion from chapters to the main body of the text. }
    \label{DepthFirstGeneration}
    %\vspace{-10pt}
\end{figure*}
% Using the methods described above, we can form tree-like structure information from the novel text in a reverse manner.
% To endow the large language model with the linguistic style of a novel and enable it to learn text expansions at different hierarchical levels, we develop five distinct categories of prompts, as shown in Figure \ref{Instructions}, to construct corpora for supervised fine-tuning. 

% Additionally, considering the special nature of the beginning and ending parts, we make distinctions when constructing the instructions. Within each expansion instruction, we incorporate preceding context information and historical knowledge about relevant entities to enhance the model's ability to make connections and maintain contextual consistency.

During the \textit{Extracting} process, we form tree-like structure information from the novel text in a reverse manner.
To endow the large language model with the linguistic style of a novel and enable it to learn text expansions at different hierarchical levels, we develop five distinct categories of prompts, as shown in Figure \ref{Instructions}, to construct corpora for supervised fine-tuning. 
These instruction formats of the corpora contain all the instruction formats required by the LLM in the \textit{Expanding} stage. The writing assistant will maintain contextual consistency through the previous outline, current outline and historical knowledge about relevant entities during the \textit{Expanding} process.

% We use approximately 800 novels to construct corpora following the process described above. We then fine-tune the open-source model BaiChuan2-13B-Chat \citep{baichuan2023baichuan2} using these corpora for 2 epochs, stopping when the loss reaches around 1.0 to prevent severe overfitting. The fine-tuned model can be employed with various instructions to implement narrative expansion and generate text with a writing style that closely aligns with the genre of novels.

We collect approximately 800 web novels from the internet, covering genres such as fantasy, suspense, romance, and modernity, to construct corpora following the process described above. We provide the distribution of the total word count and total number of chapters of these novels in Appendix \ref{dataset_info}. We then fine-tune the open-source model BaiChuan2-13B-Chat \citep{baichuan2023baichuan2} using these corpora for 2 epochs, stopping when the loss reaches around 1.0 to prevent severe overfitting. The fine-tuned model can be employed with various instructions to implement narrative expansion and generate text with a writing style that closely aligns with the genre of novels.

%% medium human evaluation
\begin{table*}[t]
\centering
\resizebox{\textwidth}{!}{
    \begin{tabular}{l c c c c c c c}
    \hline
    \textbf{\makecell{Medium Novels \\ \textasciitilde{}4k words}} & \textbf{Relevant} & \textbf{Coherent} & \textbf{Event} & \textbf{Diction} & \textbf{Transition} & \textbf{Interesting} & \textbf{Human-like}\\ 
    \hline
    Rolling-GPT & 13.34 & 6.67 & 0.00 & 10.00 & 6.67 & 6.67 & 10.00 \\
    Ex3 & \textbf{63.34} & \textbf{83.34} & \textbf{53.33} & \textbf{80.00} & \textbf{93.34} & \textbf{80.00} & \textbf{96.67} \\
    \hline
    DOC & 40.00 & 30.00 & 23.33 & 40.00 & 23.33 & 26.67 & 33.33 \\
    Ex3 & 50.00 & \textbf{66.67} & \textbf{73.33} & \textbf{66.67} & \textbf{60.00} & \textbf{46.67} & \textbf{80.00} \\
    \hline
    RE3 & 16.67 & 20.00 & 26.67 & 36.67 & 23.33 & 16.67 & 16.67 \\
    Ex3 & \textbf{63.33} & \textbf{56.67} & \textbf{53.33} & \textbf{70.00} & \textbf{53.33} & \textbf{63.33} & \textbf{73.33} \\
    \hline
    R\scalebox{0.9}{ECURRENT}GPT & 3.33 & 3.33 & 3.33 & 0.00 & 26.67 & 0.00 & 0.00 \\
    Ex3 & \textbf{46.67} & \textbf{70.00} & \textbf{66.67} & \textbf{86.67} & \textbf{80.00} & \textbf{83.33} & \textbf{100.00} \\
    \hline
    \end{tabular}
}
%\vspace{-8pt}
\caption{\label{tab:medium_human_evaluation}
Pair-wise comparison of Ex3 with different baselines for 10 novels of different genres under the scale of medium-length novels.
The horizontal axis represents different methods being compared, the vertical axis represents the metrics being compared, and the table values indicate the percentage of annotators who believe the method's novels align with the metrics.
Results in different comparisons are not comparable with each other. 
Bold indicates significance with $p<0.05$. 
}
%\vspace{-12pt}
\end{table*}

\subsection{Expanding}
To enable large-scale models to fully leverage the structure information for expansive writing and track important entity information during text generation, we employ a depth-first writing mode (Figure \ref{DepthFirstGeneration}.
In this mode, the theme and hierarchical depth of the novel are manually specified, where the theme determines the writing style, and the depth indirectly controls the final length of the generated text.  Afterward, the writing assistant output the novel title and introduction. However, users also have the option to provide their own titles and introductions if necessary. The subsequent writing process will be fully automated until a complete novel is written. During the writing process, the fine-tuned writing assistant is responsible for text expansion, while another general-purpose large language model handles entity extraction and updates the entity information database along with the text generation.

During the expansion of intermediate-level outlines, in accordance with the depth-first generation process, the expansion of the subsequent outline node can only commence after the subtree beneath the current outline node has been completely written. This meticulous approach guarantees the logical coherence of entity information within the generated outlines.

When it comes to the expansion of chapters, each chapter contains several paragraph group outlines. The writing assistant sequentially generate the main text of the chapter based on these paragraph group outlines. We utilize the outline of the current paragraph group to extract entity names for knowledge collection. After generating the main text, we use the collected knowledge and the generated text to update the entity information database.

In this depth-first tree-like generation process, the leaf nodes represent the main text of the new novel. By sequentially integrating the text, we can obtain a complete and entirely new novel that is fully generated by the writing assistant.

%% long human evaluation
\begin{table*}[t]
\centering
\resizebox{\textwidth}{!}{
    \begin{tabular}{l c c c c c c c}
    \hline
    \textbf{\makecell{Long Novels \\ \textasciitilde{}10k words}} & \textbf{Relevant} & \textbf{Coherent} & \textbf{Event} & \textbf{Diction} & \textbf{Transition} & \textbf{Interesting} & \textbf{Human-like}\\ 
    \hline
    Rolling-GPT & 0.00 & 13.79 & 3.45 & 10.35 & 6.90 & 6.90 & 17.24 \\
    Ex3 & \textbf{86.21} & \textbf{75.86} & \textbf{89.66} & \textbf{96.56} & \textbf{75.87} & \textbf{82.76} & \textbf{93.10} \\
    \hline
    DOC & 15.63 & 15.63 & 31.25 & 25.00 & 15.63 & 21.88 & 25.01 \\
    Ex3 & \textbf{65.63} & \textbf{50.00} & \textbf{53.13} & \textbf{59.38} & \textbf{53.13} & \textbf{62.50} & \textbf{65.63} \\
    \hline
    RE3 & 12.90 & 25.81 & 9.68 & 25.80 & 25.81 & 19.35 & 29.03 \\
    Ex3 & \textbf{74.19} & \textbf{51.61} & \textbf{74.20} & \textbf{74.20} & \textbf{54.84} & \textbf{64.52} & \textbf{64.52} \\
    \hline
    R\scalebox{0.9}{ECURRENT}GPT & 34.48 & 6.90 & 6.90 & 0.00 & 10.34 & 6.90 & 0.00 \\
    Ex3 & \textbf{55.17} & \textbf{86.21} & \textbf{34.48} & \textbf{82.76} & \textbf{68.97} & \textbf{68.97} & \textbf{96.55} \\
    \hline
    \end{tabular}
}
%\vspace{-8pt}
\caption{\label{tab:long_human_evaluation}
Pair-wise comparison of Ex3 with different baselines for 10 novels of different genres under the scale of long-length novels.
The horizontal axis represents different methods being compared, the vertical axis represents the metrics being compared, and the table values indicate the percentage of annotators who believe the method's novels align with the metrics.
Results in different comparisons are not comparable with each other. 
Bold indicates significance with $p<0.05$. 
}
%\vspace{-12pt}
\end{table*}

\section{Evaluation}

\paragraph{Experiment Setup.}
Our approach has the capability to generate a long-form novel by giving a short premise.
To evaluate the effectiveness of Ex3, we compare the novels generated by Ex3 and other baselines by inputting the same premise, similar to the setup in the \citet{yang2022doc}.

\paragraph{Method Instantiation}
% We select 20 high-quality premises as input for the evaluation, which are selected from manually written and real novels from outside the training datasets.
We select 20 high-quality premises as input, which are chosen from real novel data and will not appear in the training dataset.
The length of these premises ranges from 60 to 120 words, covering various genres including science fiction, fantasy, mystery, and romance.
% Our proposed approach is capable of generating very lengthy novels, a task that the previous baselines is not proficient in.
The length of the generated novels by Ex3 can be very long, which was not observed in the previous baseline experiments.
To compare as fairly as possible with previous work, we reference the novel length in the R\scalebox{0.9}{ECURRENT}GPT baseline \citep{zhou2023recurrentgpt} and design two sets of novels with different lengths for experimentation.
% For comparison,  
We generate a set of long-length novels (\textasciitilde{}10k words) and a set of medium-length novels (\textasciitilde{}4k words) respectively, where each word means a Chinese character.

\paragraph{Baselines.}
We selected the following four baselines to compare against Ex3.
\begin{enumerate}
  \item \textbf{Rolling-GPT}: A simple baseline that uses a sliding window to prompt LLM to generate novels.
  % a simple baseline that prompts GPT3-175B with a sliding window approach, directly leveraging the premise and previously generated content to generate a continuous novel until the maximum length is reached.
  \vspace{-0.5\baselineskip}
  \item \textbf{DOC}: The major baseline, based on Paper \citet{yang2022doc}, utilizes hierarchical structure to prompt LLM to transform the premise into a plan and then generate novels.
  We only modify the parameter controlling the generation length, while keeping other settings as the default configuration in the paper.
  % the main baseline, which is based on \citet{yang2022doc}, which prompts LLM to transform the premise into an outline first, then generate novels via a hierarchical structure.
  % For fair comparison, we use the default settings in their paper and only modify the parameters that control the generated length to align with the experiment.
  \vspace{-0.5\baselineskip}
  \item \textbf{RE3}: a baseline based on \citet{yang2022re3}, which also utilizes hierarchical structure to generate novels by taking a premise as input.
  Experimental settings are consistent with DOC for fair comparison.
  % Similarly, we only modify the parameter controlling the generation length of Ex3 for fair comparison.
  \vspace{-0.5\baselineskip}
  \item \textbf{R\scalebox{0.9}{ECURRENT}GPT}: a baseline based on \citet{zhou2023recurrentgpt}, which allows generating a novel step by step.
  We use the LLM to convert the premise into the format required for inputting into the R\scalebox{0.9}{ECURRENT}GPT, and only adjust the parameter controlling the generation length.
  % \vspace{-0.5\baselineskip}
\end{enumerate}
  % \item \textbf{RE3}: a baseline based on \citet{yang2022re3}, which is an early version of DOC and employs the same novel generation mechanism as DOC.
  % Similarly, we only modify the parameter controlling the length of Ex3, while reusing the other settings from the DOC baseline.

\paragraph{Human Evaluation.}
Our goal is to generate longer and more human-like novels.
% Although defining "human-like" is a complex issue, we can still assess whether the novels are more readily accepted through human evaluation.
Following prior work like \citet{yang2022doc} and \citet{zhou2023recurrentgpt}, We conduct a pairwise experiment to compare Ex3 and the baseline by designing questionnaires and presenting them to annotators for scoring.
The questionnaire consisted of a premise, two novels (denoted as A and B, with random order), and a set of questions.
The human evaluation experiment includes the following metrics:

% We employ the commonly used metrics of \textbf{Relevant}: the proportion of the novel text aligning with the corresponding premise.
% \textbf{Coherent}: the proportion of the consistency of the contextual content within the novel text.
% \textbf{Interesting}: the proportion of human readers finding the novel's content engaging.
% % In addition to the above commonly used metrics, 
% Furthermore, we also set metrics for measuring the literary quality of the novel:
% \textbf{Event}: the proportion of individuals who find the plot of the novel more captivating.
% \textbf{Diction}: the proportion of individuals who consider the prose of the novel more elegant.
% \textbf{Transition}: the proportion of individuals who consider the transition between paragraphs in the novel to be more natural.
% Finally, we also set \textbf{Human-like} to indicate the proportion of the generated novels that are considered to be closer to human-written novels.
\begin{itemize}
    \item \textit{Relevant}: Consistent with the premise.
    \vspace{-0.5\baselineskip}
    \item \textit{Coherent}: Contextual consistency.
    \vspace{-0.5\baselineskip}
    \item \textit{Interesting}: Interesting to the reader.
    \vspace{-0.5\baselineskip}
    \item \textit{Event}: More captivating plot.  
    \vspace{-0.5\baselineskip}
    \item \textit{Diction}: More elegant grammar.
    \vspace{-0.5\baselineskip}
    \item \textit{Transition}: The transition is more natural.
    \vspace{-0.5\baselineskip}
    \item \textit{Human-like}: Considered to have a style closer to human-written.
\end{itemize}
% \textbf{Relevant}.  Consistent with the premise.  

% \textbf{Coherent}. Contextual consistency.  

% \textbf{Interesting}. Interesting to the reader.

% \textbf{Event}. More captivating plot.   

% \textbf{Diction}. More elegant grammar.  

% \textbf{Transition}. The transition is more natural.

%  \textbf{Human-like}. Considered to have a style closer to human-written.

Annotators are shown a premise and two corresponding novels(one by Ex3, and one by another baseline, random order).
They are asked to assess, under different metrics, which novel is better, or if both novels are equally good or bad.
Finally, the percentage of annotators who consider the novels generated by different methods to meet the corresponding metrics under the same experimental conditions is calculated.
Each novel pairwise is evaluated by three annotators. 
In Appendix \ref{appendix:human_study_details}, we provide further details regarding the experimental questionnaire and data statistical methods used in the Human Evaluation.

\paragraph{Automation Metrics.}
% In general, it is challenging to objectively quantify the human-like quality of generated novels. 
% To the best of our knowledge, there is currently a lack of automation evaluation metrics for long-generated novels.
% Previous works either have limitations on the length and format of the generated novels involved in the evaluation or do not include automation evaluation experiments.
% Although designing automation evaluation metrics for longer-generated novels is not the core of our work, we still conduct automation evaluation experiments to gain a more comprehensive understanding of the differences between Ex3 and human-written novels.
To the best of our knowledge, previous works lack automation evaluation experiments for long novels.
While humans have a clear understanding of what constitutes a good novel, it remains challenging to use machines to make such judgments. 
In order to have a more comprehensive understanding of the differences between Ex3 and human writing, we design a set of simple automation evaluation experiments, which we think are also valuable.
\begin{figure*}[ht]
    \centering
    \resizebox{\textwidth}{!}{\includegraphics{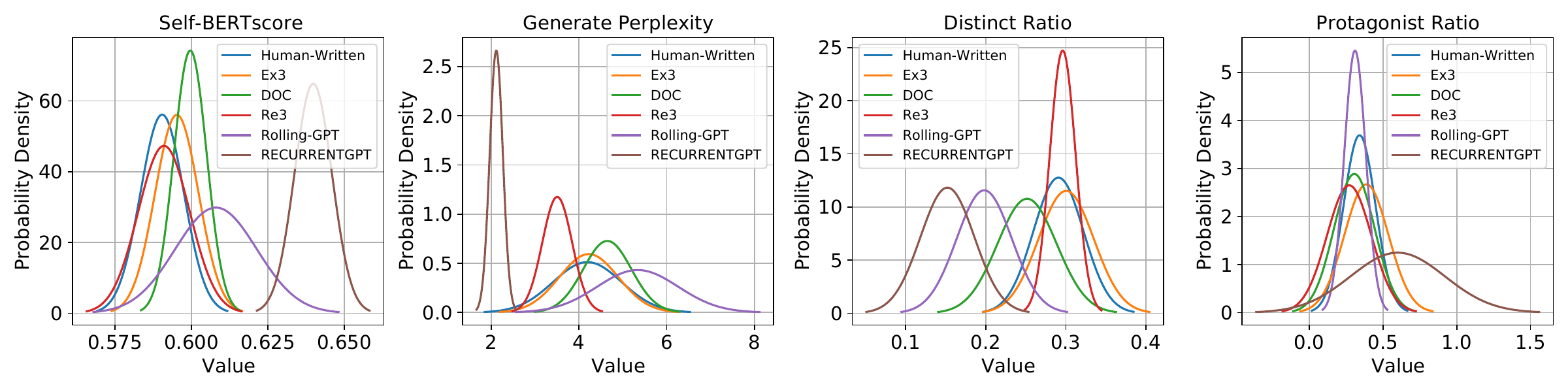}}
    \vspace{-20pt}
    \caption{The automation metric plot}
    \label{fig:automation_result}
    %\vspace{-10pt}
\end{figure*}
\begin{table*}
\centering
% medium novel table
\begin{tabular}{l c c c c c c c}
\hline
\textbf{\makecell{Method}} & \textbf{Relevant} & \textbf{Coherent} & \textbf{Event} & \textbf{Diction} & \textbf{Transition} & \textbf{Interesting} & \textbf{Human-like}\\ 
\hline
NO-TRAIN & 9.52 & 28.57 & 14.29 & 4.76 & 14.28 & 9.52 & 14.28 \\
Ex3 & \textbf{47.62} & \textbf{57.15} & \textbf{52.38} & \textbf{61.90} & \textbf{52.38} & \textbf{66.67} & \textbf{85.71} \\
\hline
NO-ENTITY & 28.57 & 28.57 & 23.81 & 28.57 & 28.57 & 38.10 & 33.33 \\
Ex3 & \textbf{52.38} & \textbf{66.67} & \textbf{52.38} & 28.57 & \textbf{66.67} & 28.57 & \textbf{61.90} \\
\hline
\end{tabular}
%\vspace{-8pt}
\caption{\label{tab:ablation_evaluation}
Pair-wise comparison of Ex3 with ablations without fine-tuned LLM and entity extraction module, respectively, for 10 novels of different genres.
The horizontal axis of represents different methods being compared, the vertical axis represents the metrics being compared, and the table values indicate the percentage of annotators who believe the method's novels align with the metrics.
Results in different comparisons are not comparable with each other. 
Bold indicates significance with $p<0.05$. 
}
%\vspace{-14pt}
\end{table*}
We select 10 premises from the human evaluation experiments and compare the distribution of different automation metrics between the human-written novels and generated novels. %corresponding to these premises.
The automation metrics are as follows:
\begin{itemize}
  \vspace{-0.25\baselineskip}
  \item \textit{Self-BERTscore}: Calculate the BERTscore \citep{zhang2019bertscore} between each adjacent pair of sentences within the novel and take the average of all scores greater than 0.5.
  % BERTscore \citep{zhang2019bertscore} can be used to calculate the similarity between two text. 
  % "Self" refers to computing the BERTscore between every adjacent pair of sentences within a novel and taking the average of all scores greater than 0.5. 
  % We consider that if the BERTscore between two sentences is less than or equal to 0.5, it indicates that they belong to different paragraphs.
  \vspace{-0.5\baselineskip}
  \item \textit{Generation Perplexity}: Average perplexity of all sentences in a novel calculated by Bert-Base-Chinese model, which represents the average syntactic level of the novel's sentences.
  % Perplexity is a commonly used metric to measure the quality of the generated text.
  % In this study, we calculated the average perplexity of all sentences in a novel by Bert-Base-Chinese model, which represents the average syntactic level of the novel's sentences.
  \vspace{-0.5\baselineskip}
  \item \textit{Distinct Ratio}: Percentage of overall vocabulary to the number of tokens in novels as a distinct ratio, following \citet{han2022go}.
  % In order to gain further insights into the differences in vocabulary usage between generated novels and human-written novels, we calculate the percentage of overall vocabulary to the number of tokens in novels as a distinct ratio, following \citet{han2022go}.
  \vspace{-0.5\baselineskip}
  \item \textit{Protagonist Ratio}: Ratio of roles in the novel that appear more than a specific threshold number of times.
  % One of the most important elements in a novel is roles.
  % We calculate the ratio of roles in the novel that appear more than a specific threshold number of times.
  % We define "valid roles" as those who appear with a frequency above a certain threshold, and then calculate the percentage of valid roles in relation to all roles.
  \vspace{-0.5\baselineskip}
\end{itemize}

% We then calculate the similarity measures between the reference novels and the generated novels based on different metrics.
% By comparing the distribution differences between different methods and human-authored novels, we determined which methods exhibited a higher degree of human-likeness.
% We take the original text corresponding to the premise as the ground truth, report bias of \textbf{Self Bert Score}: bert-score between each sentence and its preceding, \textbf{Generation Perplexity}: generated novels’ perplexity, \textbf{Type Token Ratio}: unique token ratio and \textbf{Protagonist Ratio}: ratio of effective appearance roles.
%% automation subplot

\paragraph{Results.}
As shown in Table \ref{tab:medium_human_evaluation}, we find that under the scale of medium-length novels, human readers consider Ex3 to be better consistent with the premise and also prefer Ex3's plot development and literary devices compared to other baselines. In addition, they believe that the novels produced by Ex3 are more interesting and human-like than the other methods.
This is in line with our intuition that Ex3 can learn the plan and linguistic capacity from raw novel data and produce better novels from every aspect.
Table \ref{tab:long_human_evaluation} shows the human evaluation results for the scalability of long novel generation.
It can be observed that Ex3 maintains a significant advantage over other baselines under different human metrics, indicating the potential of Ex3 for generating better longer novels. We also compare the novels generated by Ex3 with human-writing novels and provide the results in Appendix \ref{vs_human}.

% The human evaluation experiment only compared the differences between Ex3 and different baseline methods for novel generation, while the automatic evaluation experiment demonstrated the differences between different methods and human novels, as shown in Figure \ref{fig:automation_result}.
Figure \ref{fig:automation_result} shows the automation evaluation results.
It can be seen that the distribution of Ex3 under different automation metrics is comparable to that of human-written novels, particularly under the metrics of Generate Perplexity and Distinct Ratio. %, Ex3 demonstrates the highest similarity to human-written novels.
The results of the automation evaluation are consistent with our ideas that fine-tuning LLM with human-written novels can enhance the human-like quality of generated content.

\section{Ablation Study}

\paragraph{Ablated Modules.}
In order to better understand the effect of different components in Ex3 on the quality of the generated novels, we set up ablation experiments for different components.
The ablation experiment for the setup is shown below:

\begin{enumerate}
  \vspace{-0.25\baselineskip}
  \item \textsc{NO-TRAIN}: A version that skips the extracting and excelsior stages and uses the original BaiChuan2-13B-Chat model to generate novels.
  % In order to align with the original format during the generation process, we modify the prompts of the NO-TRAIN version, which do not impact the generated content.
  \vspace{-0.5\baselineskip}
  \item \textsc{NO-ENTITY}: A version that does not use the entity extraction module to generate novels.
  % We set the entity information in the prompt as "None" to block the module. %on the generation of novels.
  \vspace{-0.5\baselineskip}
\end{enumerate}

\paragraph{Results.}
As shown in Table \ref{tab:ablation_evaluation}, Ex3 exhibits significant advantages over NO-TRAIN across all metrics, indicating the critical contribution of the fine-tuned LLM to the generation of long novels that are well-liked by human readers.
In comparison to NO-ENTITY, although Ex3 performs relatively equally or slightly inferior in terms of Diction and Interesting metrics, it still maintains significant advantages in other metrics, suggesting a positive impact of the entity extraction module on enhancing the coherence of roles plots in generated novels. we also examine the output of the intermediate steps in our method and provide the results in Appendix \ref{inter_res}.
% We speculate that the reason for the entity extraction module's diminishing effect on the interesting aspects of the novels is its constraint on the generated content of the LLM, thereby limiting the full expression of its creative capabilities.

\section{Conclusion}
In this paper, we present Ex3, an automatic novel writing framework.
Ex3 first automatically extracts structure information by learning from raw novel data and then creates an instruction-following dataset with this structure information and the novel data. Subsequently, this dataset is employed to fine-tune the LLM for excelsior generation performance. With the tuned LLM, Ex3 employs a tree-like expansion method to generate 
arbitrarily long novels.
The experiment results demonstrate that Ex3 is able to produce higher-quality long-form novels than previous methods.

\section{Acknowledgement}
The authors would like to thank Meng Yuan, from Cambricon Technologies for her helpful discussions and her inspiration and contribution to the data collection.
This work is partially supported by the NSF of China (under Grants U20A20227, U22A2028, 62102399, 61925208, 62222214, 62341411, 62102398, 62372436, 62302478, 62302482, 62302483, 62302480), Strategic Priority Research Program of the Chinese Academy of Sciences, (Grant No. XDB0660200, XDB0660201, XDB0660202, CAS Project for Young Scientists in Basic Research (YSBR-029), Youth Innovation Promotion Association CAS and Xplore Prize.
\section*{Limitations}
Our proposed framework, Ex3, effectively learns novel structures and language styles from raw novel data, enabling the generation of sufficiently long novels across various genres. In our experiments, we compared Ex3 with other mainstream long-novel generation frameworks and found it to outperform them on multiple metrics. However, Ex3 also has some limitations.

While Ex3 controls the length of the generated text by adjusting the number of hierarchy levels, we were unable to comprehensively evaluate its performance on longer novels due to time and resource constraints. Additionally, our training and generation primarily focused on Chinese novels, and further testing is needed to assess the performance of Ex3 on novels in other languages. In terms of the mechanism process, the initial outline generated may have some shortcomings, and introducing an auditing and revision mechanism could enhance the coherence and brilliance of the novels. It is worth mentioning that in the \textbf{Expanding} process, the expansion of the outline is fully transparent and modifiable for the user, allowing for interactive generation. However, we did not explicitly mention this in the main text.

To address these limitations, our future work will focus on optimizing the mechanism, expanding to multiple languages, and exploring interactive generation modes to further enhance our research.

\section*{Ethics Statement}
Large language models can produce false and harmful content. In this article, we use audited human novel data to fine-tune the LLM, reducing the potential for harmful content. In addition, abuse of the automatic novel generation system can also have a bad effect on the field of novel writing, so the application of the system may require censorship and restraint.


% \bibliographystyle{acl_natbib}
% \bibliography{acl2023}
\newpage
\appendix
\section{Details on Extract Stage}
In the \textbf{Extract} stage, we use similarity to measure the relationship between texts, with the similarity calculation performed using the text2vec-base-chinese-paraphrase model \citep{text2vec}. Additionally, we construct hierarchical structure information by summarising, and employ entity extraction methods to track the development of important characters in real time. For these two tasks, we have chosen the BaiChuan2-13B-Chat \citep{baichuan2023baichuan2}, which has shown outstanding performance in Chinese. We have designed appropriate prompts to automate these two processes.

In the \textbf{Excelsior} stage, we have designed several distinct prompts to differentiate information across different levels.

\textbf{Main Text}: "Please provide a brief summary for the following paragraph group plot:{part of the main text}";

\textbf{Chapters}: "Please provide a brief summary for the following chapter plot, trying to cover the overall plot:{the text obtained by integrating the summaries of each text group in the chapter}";

\textbf{Levels}: "Please summarize the main plot of the following content in one paragraph, trying to cover the overall plot:{the text obtained by aggregating the outline group of the previous level}";

\textbf{Novel}: "Please provide a summary of the main story plot for the following content, trying to cover the overall plot:{the text to be integrated into a novel summary obtained by Recursively Summarizing method}".

\section{Details on Excelsior Stage} 
For fine-tuning the LLM, we use an open-source pytorch framework \citep{llama-factory}. As for devices, we use four nodes, each with eight NVIDIA A100 80G. We fine-tune the model with the batch size of 256, the learning rate of 5e-5, the weight decay of 1e-4 for 2 epoch.
\section{Details on Expand Stage}
We have designed different prompts for different levels to differentiate them, as illustrated in \textbf{Excelsior}. In this section, we take the generation process of a novel by setting the level depth to 1 to provide detailed prompts and outputs for each process.

\textbf{Novel Title}:Wudang Sword Immortal Song Qingshu

\textbf{Novel Introduction}: The story begins with Song Qingshu's time travel to the Blue Star. He becomes a disciple at Wudang Mountain and excels in martial arts, becoming a formidable warrior. As he resolves conflicts among the martial arts communities, Song Qingshu gradually finds himself entangled in a web of grudges and disputes. Faced with the temptation of demonic influences, he chooses to relinquish the Forbidden Evil Sword Manual and joins forces with the sect leader Zhang Sanfeng to confront the Shaolin monks. Meanwhile, the encounter between Huashan disciple Linghu Chong and Hengshan disciple Yi Lin further complicates the martial world relations.

\subsection{First Outline Expansion}
\paragraph{Prompt} \textit{You are now a writer of martial arts, fantasy, and mystery novels, currently working on a book titled "Sword Immortal Song Qingshu of Wudang." Based on the novel introduction provided, please create an outline for the story while ensuring the coherence of the plot.} \textbf{Novel Introduction}: \textit{The story begins with Song Qingshu's time travel to the Blue Star. He becomes a disciple at Wudang Mountain and excels in martial arts, becoming a formidable warrior. As he resolves conflicts among the martial arts communities, Song Qingshu gradually finds himself entangled in a web of grudges and disputes. Faced with the temptation of demonic influences, he chooses to relinquish the Forbidden Evil Sword Manual and joins forces with the sect leader Zhang Sanfeng to confront the Shaolin monks. Meanwhile, the encounter between Huashan disciple Linghu Chong and Hengshan disciple Yi Lin further complicates the martial world relations.}

\paragraph{Output}
\textbf{Part 1}: This part tells the story of Song Qingshu's journey to the Blue Star, his apprenticeship and training at Wudang Mountain, and his involvement in the martial arts conflicts. Throughout this process, Song Qingshu grows and becomes a martial arts master, resolving conflicts within the martial world. However, he also gets entangled in various grudges, including conflicts with his master, a love affair with a woman in red, and battles with martial artists. When faced with the temptation of demonic music, Song Qingshu chooses to give up the Forbidden Evil Sword manual and joins his master Zhang Sanfeng in opposing the Shaolin monks, thus upholding martial peace.

\textbf{Part 2}: This part tells the story of Huashan disciple Linghu Chong's encounter with Hengshan disciple Yi Lin and their adventures in the martial world. Meanwhile, the conflict between Wudang disciple Song Qingshu and Dongfang Bubai continues. Throughout this process, the shifting powers within the martial world, complex relationships between characters, and entangled grudges of the jianghu are depicted. Song Qingshu emerges victorious in his duel with Dongfang Bubai, revealing the truth about Lin Yuan's story and also disclosing Duan Yu's identity. In this world full of jianghu grudges, various heroes make their appearances, showcasing the true essence of martial arts spirit and the demeanor of a true gentleman.

\textbf{Part 3}: Wudang disciple Song Qingshu interacted with various characters and experienced a series of adventures and challenges. He constantly improved his cultivation from his conflict with Xiao Yunling, who disguised himself as a man, to his friendship with the mysterious woman Mu Renqing. During this process, he also made friends such as Mu Wanqing to complete tasks together, resist foreign enemies, and defend the peace of the martial arts world. Song Qingshu conflicts with Shaolin Temple on the issues of justice and evil, demonstrating strong strength and unique personality. Song Qingshu went to Lingjiu Temple to answer the monk's question and rescue Mu Wanqing, but was enlightened by the abbot. Finally, she saved Mu Wanqing and spent the rest of her life with her.

The times of outline expansions depend on the design of the depth level. When the depth level is set to 1, outline expansion will occur only once. However, when set to a higher depth level, there will be additional outline expansions in subsequent steps.

\subsection{Chapter Arrangement of Part 1 (Opening Part)}
\paragraph{Prompt} 
\textit{You are currently a martial arts, fantasy, and mystery novelist working on a book titled "Song Qingshu, the Sword Immortal of Wudang." Building upon the previous storylines, historical information about characters or locations is as follows: None.} \textbf{Summary}: \textit{This part tells the story of Song Qingshu's journey to the Blue Star, his apprenticeship and training at Wudang Mountain, and his involvement in the conflicts of the martial world. Throughout this process, Song Qingshu continuously grows and becomes a formidable martial artist, resolving conflicts through his interactions with martial artists. However, he also becomes entangled in various grudges, including conflicts with his master, a love affair with a woman in red, and battles with martial artists. When faced with the temptation of demonic music, Song Qingshu chooses to abandon the Forbidden Evil Sword manual and joins his sect leader Zhang Sanfeng in opposing the Shaolin monks, thus upholding martial peace.}

\paragraph{Output}
\textbf{Chapter 1}: Song Qingshu travels to a planet called the Blue Star and undergoes a series of adventures before becoming an apprentice at Wudang Mountain. At Wudang Mountain, he gains attention due to a dragon turtle egg and learns Taoist spells and martial arts from his master. During this time, he also encounters a mysterious woman, and they fall in love. Song Qingshu cultivates at Wudang Mountain, mastering the essence of Taoism, and becomes a renowned martial artist.

\textbf{Chapter 2}: As a young man, Song Qingshu joins the Three Emperor Cannon Fist Sect in search of the mysterious woman, and through enduring hardships, he becomes an exceptional martial artist. Along the way, he befriends He Tianhua and together they overcome challenges and resolve grudges. Song Qingshu becomes a respected hero in the martial world.

\textbf{Chapter 3}: Song Qingshu becomes entangled in a series of complex martial conflicts, ...

\subsection{Chapter Arrangement of Part 3 (Ending Part)}
\paragraph{Prompt}
\textit{As a martial arts, fantasy, and mystery novelist, you are currently writing a novel titled "Song Qingshu, the Sword Immortal of Wudang." Building upon the previous storylines and historical information of relevant characters and locations, here's an expanded plot based on the provided summary.} \textbf{Previous Storylines}:\textit{
The encounters between Huashan Sect disciple Linghu Chong, Hengshan Sect female disciple Yi Lin, and their adventures in the martial world are depicted. The conflicts between Wudang disciple Song Qingshu and Dongfang Bubai are recurring themes. Throughout these events, the shifting power dynamics in the martial arts world, intricate character relationships, and entanglements of Jianghu grievances are showcased. Song Qingshu emerges victorious in his duel against Dongfang Bubai, unraveling the story of Lin Yuantu while also revealing Duan Yu's true identity. In this world full of Jianghu feuds, various heroes make their appearances, embodying the true essence of martial arts and the demeanor of a true gentleman.} \textbf{Historical Information of Characters and Locations}: \textit{
\textbf{Song Qingshu} | A highly accomplished cultivator; he rescued the village girl Liu Susu from bandits and developed feelings for her before being forced to separate. He has a connection with the villagers below Wudang Mountain, particularly Huang Ziyu.
\textbf{Longmen Ghost King} | Sworn brother of Yi Lin's master.
\textbf{Wudang Mountain} | A sacred place of Taoism, where Song Qingshu, Zhang Songxi, and other characters engage in activities. It is also the site of conflicts between Jiao Zhishang and Lan Ruozhi.
\textbf{Duan Yu} | Prince of Dali Kingdom, who initially meets Song Qingshu and engages in a conversation. He later catches the desperate Longmen Ghost King, who is Duan Yu's adversary. There may be a connection between Duan Yu and Dongfang Bubai.
\textbf{Linghu Chong} | Disciple of Huashan Sect, who encounters Yi Lin and discusses her identity and purpose of seeking him.
\textbf{Yi Lin} | Female disciple of Huashan Sect, who meets Linghu Chong and reveals that she is seeking her companion.} 
\textbf{Outline} \textit{Song Qingshu, as a disciple of Wudang, interacts with various characters, experiencing a series of adventures and challenges. From his conflicts with the cross-dressing Xiao Yunling to his friendship with the mysterious woman Mu Renqing, he continuously enhances his cultivation. Along the way, he also befriends individuals like Mu Wanqing, collaborating to complete missions, fend off external threats, and defend the peace of the martial world. Song Qingshu engages in a clash with Shaolin Temple over matters of justice and evil, showcasing his formidable strength and unique personality. To answer a monk's question and rescue Mu Wanqing, he embarks on a journey to the Lingjiu Temple, where he is enlightened by the abbot. Ultimately, he saves Mu Wanqing and spends the rest of his life together with her.}

\paragraph{Output}
\textbf{Chapter 8}: Wudang disciple Song Qingshu interacts with various characters, embarking on numerous adventures and facing challenges. He clashes with Xiao Yunling, a woman disguised as a man, and discovers her true identity while searching for her. Simultaneously, he enhances his cultivation through his interactions with a mysterious woman. Throughout this journey, he also befriends many individuals, such as Mu Wanqing, and together they experience battles and challenges. Song Qingshu successfully improves his own cultivation and demonstrates his formidable strength in combat against enemies.

\textbf{Chapter 9}: This is a story of martial arts conflicts, love, and hatred. Wudang disciple Song Qingshu becomes entangled in a struggle involving major sects of the martial world, confronting the Lingjiu Temple, the Demon Sect, and other forces. Meanwhile, his relationship with his childhood sweetheart, Mu Wanqing, gradually intensifies. In the midst of this turmoil, Song Qingshu showcases his leadership abilities and sense of justice, striving to protect the peace of the martial world. However, he also faces challenges from the Shaolin Temple, particularly regarding the question of justice and evil posed by the eminent monk Yuan Kong of the Lingjiu Temple. Song Qingshu must provide a satisfactory answer within seven days, or else face punishment. Throughout this storyline, Song Qingshu displays his intelligence, courage, and determination, earning the respect of many. In terms of his emotions, he and Mu Wanqing also endure many hardships and decide to spend their lives together.

\subsection{Plot Arrangement of Chapter 1 (Opening Chapter)}
\paragraph{Prompt} \textit{As a martial arts, fantasy, and mystery novel writer, you are writing the opening plot of a novel called "Wudang Sword Immortal Song Qingshu". Please expand the opening plot according to the outline of the opening chapter.} \textbf{Opening chapter outline}: \textit{Song Qingshu traveled to a planet called Blue Star and after experiencing a series of adventures, he apprenticed at Wudang Mountain. In Wudang Mountain, he attracted attention to a dragon turtle egg and learned Taoist spells and martial arts from his master. At the same time, he also met a mysterious woman, and the two fell in love. Song Qingshu practiced at Wudang Mountain, mastered the essence of Taoism, and became a generation of martial arts masters.}

\paragraph{Output}
\textbf{Plot 1}: 
A young man named Song Qingshu traveled to a planet called Blue Star, where he experienced many adventures and challenges. By chance, he obtained a mysterious dragon turtle egg, which brought about a revolutionary change in his life. He was adopted as a disciple by a Taoist from Wudang Mountain and began his path of cultivation. Song Qingshu practiced at Wudang Mountain and gradually mastered the essence of Taoism. During this process, he experienced many wars and adventures, demonstrating his bravery and wisdom.

\textbf{Plot 2}:
Song Qingshu studied Taoism under his master at Wudang Mountain and lived with his master and fellow disciples. One day, he discovered that dragon turtle by the lake and after sharing it with his master, he decided to live alone. During his solitary residence, Song Qingshu was restless and later sought help from his master, hoping to take him on a trip to the mortal world. The master agreed to his request, but requested that he return to Wudang Mountain to continue his cultivation.

\textbf{Plot 3}:
Song Qingshu studied Taoist magic with his master at Wudang Mountain and gradually grew into a master. One day, while practicing the Five Thunder Correct Method, he unexpectedly caught the attention of a girl named Huang Ziyu. Huang Ziyu was surprised by Song Qingshu's body movements, while Song Qingshu was attracted by her beauty. Song Qingshu pursued Huang Ziyu and learned martial arts from her. The two spent a wonderful time together in Wudang Mountain.

\subsection{Plot Arrangement of Chapter 9 (Ending Chapter)}
\paragraph{Prompt} \textit{As a writer of martial arts, fantasy, and mystery novels, you need to wrap up the story of "Sword Immortal Song Qingshu of Wudang." The provided outline includes the synopsis of previous chapters, historical information about relevant characters or locations, and an outline for the concluding chapter. Now, let's expand on the plot for the final chapter.} \textbf{Previous Chapter Outline}: \textit{Wudang disciple Song Qingshu encounters obstacles from various major sects and goes through numerous adventures and challenges. He clashes with Xiao Yunling, who disguises herself as a man, and during his search for her true identity, he discovers her secret. Meanwhile, he also enhances his cultivation through interactions with a mysterious woman. Throughout this journey, he coincidentally reunites with his childhood sweetheart, Mu Wanqing, during a mission to punish evildoers. Together, they face numerous battles and challenges. Song Qingshu successfully improves his cultivation and showcases his formidable strength in battles against enemies.} \textbf{Historical Information about Characters or Locations}: \textit{\textbf{Song Qingshu} | A highly skilled martial artist; rescued his hometown girl, Liu Susu, from bandits. Developed feelings for Liu Susu but faced numerous obstacles, leading to their separation. Has a friendly relationship with the villagers at the foot of Huangzi Mountain. Childhood friends with Mu Wanqing and reunites with her during a chance encounter in a fight.
\textbf{Mu Wanqing} | Song Qingshu's childhood sweetheart; admires Song Qingshu.\textbf{Duan Yu} | Prince of Dali Kingdom; initially meets Song Qingshu and they exchange insights. Later, Duan Yu saves Song Qingshu from a dire situation involving the Ghost King of Dragon Gate.
\textbf{Linghu Chong} | Disciple of Huashan sect; meets Yi Lin and may have a connection with Dongfang Bubai.
\textbf{Yi Lin} | Female disciple of Huashan sect; meets Linghu Chong and expresses her purpose of finding companions.}\textbf{Concluding Chapter Outline}:\textit{
Wudang disciple Song Qingshu became embroiled in a conflict involving major martial sects, including Lingjiu Temple and the Demon Cult. Simultaneously, his romantic relationship with childhood sweetheart Mu Wanqing gradually deepens. In the midst of this turmoil, Song Qingshu demonstrates his leadership abilities and sense of justice, striving to protect peace in the martial world. However, he also faces a challenge from the Shaolin Temple, particularly regarding the question of justice and evil raised by the eminent monk Yuan Kong. Song Qingshu must provide a satisfactory answer within seven days, or face punishment. He displays his wisdom, courage, and determination, earning the respect of those around him. In terms of his romantic relationship, he and Mu Wanqing face various hardships but ultimately decide to spend their lives together.}

\paragraph{Output}
\textbf{Plot 42} This section tells the story of Wudang disciple Song Qingshu and his childhood sweetheart, Mu Wanqing, as well as the events involving his fellow disciples. Due to a misunderstanding, Song Qingshu is falsely accused of murder and escapes with Zhang Long. Along the way, they encounter figures like the Golden Snake King, who reveals the truth and informs them that Mu Wanqing is being pursued. As they search for Mu Wanqing, their feelings for each other deepen. Meanwhile, the forces of good and evil engage in a fierce battle for the "Star Finger Technique," and Song Qingshu strives to protect the peace of the martial world.

\textbf{Plot 43} A group of martial artists finds themselves in a dilemma amidst the conflict between the Demon Cult and Lingjiu Temple. At this moment, Wudang disciple Song Qingshu proposes leading his fellow disciples to maintain order. With an increasing number of martial artists joining them, the situation gradually comes under control. However, the Jia brothers from the Demon Cult take advantage of the situation and escalate the conflict further. In a critical moment, Song Qingshu decisively takes action, subdues the Jia brothers, and quells the turmoil. This act garners admiration from the onlookers and elevates his status in the martial world. Subsequently, the eminent monk Yuan Kong from Lingjiu Temple poses a question that sparks a discussion about justice and evil. Song Qingshu is tasked with providing an answer, which will determine his fate.

\textbf{Plot 44} Facing a difficult decision, Song Qingshu must decide to provide a satisfactory answer within seven days. The answer to this question is intertwined with the life of his childhood sweetheart, Mu Wanqing, and the deadline marks a life-and-death moment. Song Qingshu embarks on a journey alone to Lingjiu Temple to find the answer to the question. Along the way, he coincidentally encounters Linghu Chong and engages in a martial arts contest. The abbot intervenes just in time, taking control of the situation and offering guidance to Song Qingshu. Ultimately, Song Qingshu gives his response, saving Mu Wanqing's life but at the cost of 80\% of his cultivation power. Touched by his sacrifice, Mu Wanqing decides to spend the rest of her life with Song Qingshu.

\subsection{The Main Writing of the Plot 1 (Opening Plot)}
\paragraph{Prompt} \textit{You are writing the opening paragraph for your martial arts, immortal chivalry, and suspense novel "Wudang Sword Immortal Song Qingshu". Please expand the opening content from a third-person perspective based on the plot outline of the opening paragraph.} \textbf{Plot Outline}: \textit{A young man named Song Qingshu traveled to a planet called Blue Star, where he experienced many adventures and challenges. By chance, he obtained a mysterious dragon turtle egg, which brought about a revolutionary change in his life. He was adopted as a disciple by a Taoist from Wudang Mountain and began his path of cultivation. Song Qingshu practiced at Wudang Mountain and gradually mastered the essence of Taoism. During this process, he experienced many wars and adventures, demonstrating his bravery and wisdom.}

\paragraph{Output}
Perhaps, this is fate!

...

Blue Star is a mysterious and beautiful planet. It is 120 light-years away from Earth. Here, the four seasons are distinct, the scenery is beautiful, the resources are abundant, and there are completely different biological species from the Earth.

On the Blue Star, there is a large lake called Bailang Lake. Surrounded by mountains and lush trees, the lake is surrounded by tens of thousands of acres of fertile land at the foot of the mountains. The water in the lake is clear and sandy white, like a huge mirror at the same time. There are many waterfowl in the lake, leisurely wandering on the water surface.

On one side of the lake, there is a small island called Longgui Island. The island is shaded by green trees, with blooming flowers. In the center of the island, there is an ancient well, which is said to be the oldest ancient tree on the island.
This ancient well is said to have gone through millions of years. There is a stone tablet next to the well with three lines of characters engraved on it, but it is already blurry and unrecognizable.

At noon on this day, the sun was hanging high in the sky. After a sudden loud bang, an object fell straight down from the sky, passing through the dense forest and crashing into the grass by the ancient well with a loud bang.

A large pit was dug up in the grass, and something lay at the bottom of the pit.

In this time and space, another person fell, in his thirties, tall and majestic, with thick eyebrows and big eyes, beards on his cheeks, and wearing a blue Taoist robe. The person fell to the bottom of the pit, looked up at the sky and murmured to himself, "Finally back, ten years! Ten years!" Then he hugged his head and rolled painfully on the ground.
...

\subsection{The Main Writing of the Plot 2 (Middle-Part Plot)}
\paragraph{Prompt} \textit{You are currently writing the main body of the martial arts, immortal, and suspense novel "Wudang Sword Immortal Song Qingshu". Please contact the previous summary and historical information about the characters or locations, and expand the paragraph content from a third-person perspective based on this abstract.} \textbf{Previous summary}: \textit{A young man named Song Qingshu traveled to a planet called Blue Star, where he experienced many adventures and challenges. By chance, he obtained a mysterious dragon turtle egg, which brought about a revolutionary change in his life. He was adopted as a disciple by a Taoist from Wudang Mountain and began his path of cultivation. Song Qingshu practiced at Wudang Mountain and gradually mastered the essence of Taoism. During this process, he experienced many wars and adventures, demonstrating his bravery and wisdom.} \textbf{Historical information of individuals or locations}: \textit{\textbf{Song Qingshu} | A master of successful cultivation; Rescue the young girl Liu Susu who was abducted by bandits. \textbf{Master} | Master of Wudang Mountain, teaching Song Qing the Taoist teachings. \textbf{Dragon Turtle} | A divine creature that left behind a dragon egg. \textbf{Liu Susu} | A kidnapped girl from her hometown.} \textbf{Summary of this paragraph}: \textit{Song Qingshu studied Taoism under his master at Wudang Mountain and lived with his master and fellow disciples. One day, he discovered the dragon turtle by the lake and after sharing it with his master, he decided to live alone. During his solitary residence, Song Qingshu was restless and later sought help from his master, hoping to take him on a trip to the mortal world. The master agreed to his request, but requested that he return to Wudang Mountain to continue his cultivation.}

\paragraph{Output}

The Taoist nodded and smiled, but remained silent, drifting straight up the mountain.

Upon learning that his master had agreed, Song Qingshu was overjoyed and chased after him closely. The two of them arrived at Nanyan Temple on Tianzhu Peak in Wudang Mountain, one in front of the other. I saw a red-faced old Taoist sitting on a stone stool in front of the thatched cottage, enjoying the cool breeze. Three young Taoist priests stood beside him, one with a black face and two with a white face. Seeing their master and disciple arrive, they quickly welcomed them in together.

"Master, Master." The old Taoist with a red face smiled and said to the middle-aged Taoist, "I have brought my disciple back today!"

The middle-aged Taoist nodded and said to the three young Taoists, "Have you ever seen your senior brothers? You will become fellow disciples from now on."

The three of them said in unison, "I've seen them, I've seen them"

The middle-aged Taoist said to Song Qingshu, "This is your eldest senior brother Zhu Gaotian, this is your second senior brother Jin Shiqi, and this is your third senior brother Baiyun Chan." He pointed to Song Qingshu with a black face and said, "Your senior brother's surname is Zhu, you can call him Zhu Senior Brother. Your second senior brother's surname is Jin, you can call him Jin Senior Brother, and this is your third senior brother. You can call him Bai Senior Brother. I am a relative of your master, you can call me Master Zhao."

...

\subsection{The Main Writing of the Plot 44 (Ending Plot)}
\paragraph{Prompt} \textit{The martial arts, immortal, and suspense novel "Wudang Sword Immortal: Song Qingshu" is coming to its climax. The following content provides a summary of the previous events, historical information about characters or locations, and a summary of the ending paragraph. } \textbf{Summary of Previous Events}:\textit{A group of martial artists finds themselves in a dilemma amidst the conflict between the Demon Cult and Lingjiu Temple. At this moment, Wudang disciple Song Qingshu proposes leading his fellow disciples to maintain order. With an increasing number of martial artists joining them, the situation gradually comes under control. However, the Jia brothers from the Demon Cult take advantage of the situation and escalate the conflict further. In a critical moment, Song Qingshu decisively takes action, subdues the Jia brothers, and quells the turmoil. This act garners admiration from the onlookers and elevates his status in the martial world. Subsequently, the eminent monk Yuan Kong from Lingjiu Temple poses a question that sparks a discussion about justice and evil. Song Qingshu is tasked with providing an answer, which will determine his fate.} \textbf{Historical Information about Characters or Locations}: \textit{
\textbf{Song Qingshu} | A highly skilled martial artist; rescued his hometown sweetheart Liu Susu from bandits and developed romantic feelings for her but faced numerous obstacles that led to their separation. Has a connection with the villagers beneath Huangzi Mountain. Childhood sweetheart with Mu Wanqing, and their feelings deepen during a fight over the "Star Finger Technique." Subdued the Jia brothers during the conflict between the Demon Cult and Lingjiu Temple.
\textbf{Mu Wanqing} | Childhood sweetheart of Song Qingshu; develops deeper feelings for him during the struggle for the "Star Finger Technique," secretly suffering from a chronic and deadly poison unknown to others.
\textbf{Jia Brothers} | Cause trouble during the conflict between the Demon Cult and Lingjiu Temple; subdued by Song Qingshu.
\textbf{Abbot Yuan Kong} | An eminent monk from Lingjiu Temple; reveals Mu Wanqing's secret.
\textbf{Duan Yu} | Prince of Dali kingdom; meets Song Qingshu and engages in mutual exchanges, later catches the distressed Golden Snake King.
\textbf{Linghu Chong} | Disciple of Huashan Sect, meets Yi Lin, potentially related to Dongfang Bubai.
\textbf{Zhang Long} | Disciple of Tianxing Sect, has connections with Mu Wanqing.}
\textbf{Ending Summary}:\textit{Facing a difficult decision, Song Qingshu must decide to provide a satisfactory answer within seven days. The answer to this question is intertwined with the life of his childhood sweetheart, Mu Wanqing, and the deadline marks a life-and-death moment. Song Qingshu embarks on a journey alone to Lingjiu Temple to find the answer to the question. Along the way, he coincidentally encounters Linghu Chong and engages in a martial arts contest. The abbot intervenes just in time, taking control of the situation and offering guidance to Song Qingshu. Ultimately, Song Qingshu gives his response, saving Mu Wanqing's life but at the cost of 80\% of his cultivation power. Touched by his sacrifice, Mu Wanqing decides to spend the rest of her life with Song Qingshu.}

\paragraph{Output} 
"After seven days, will you give me an answer?" 

"Yes!" came the unanimous response.

"What if I choose not to answer?" Song Qingshu inquired.

The eminent monk replied, "Then there will be no mercy, and you shall be killed!"

His words caused a stir among the attendees.

"Eminent monk, can you truly carry out what you say?" Song Qingshu questioned.

The eminent monk sternly replied, "As a monk, I do not speak falsehoods!"

Song Qingshu smiled and said, "Very well, it's a deal!"

The eminent monk declared, "After seven days, it will be a matter of life and death. If nobody can provide an answer, their heads will be taken."

Song Qingshu nodded in agreement, and the eminent monk clasped his hands together, saying, "Well done, Song benefactor. I shall take my leave."

...

After the life-or-death decision, Mu Wanqing's health gradually recovered, while Song Qingshu's cultivation suffered. However, his heart grew even more resolute. They chose to embrace a life far away from the worldly chaos and discovered a secluded valley where they decided to settle down and begin a new life.

Within the valley, the seasons were like perpetual spring, with flowers blooming in abundance and a murmuring stream, creating an otherworldly paradise.

On a moonlit night, Song Qingshu and Mu Wanqing sat in front of their bamboo house, gazing at the starry sky while holding hands. Mu Wanqing leaned on Song Qingshu's shoulder and whispered, "This kind of life is what I've always wanted."

Song Qingshu smiled and tightly grasped her hand, saying, "As long as I'm with you, whether in the martial world or in this valley, it's where my heart belongs."

Under the starlight, their shadows stretched long, very long, like the enduring love between the two. This journey was filled with hardships and trials, but it ultimately transformed into a serene and tranquil existence.[END]

\section{Statistics of Dataset Lengths}
\label{dataset_info}
\begin{figure}[!h]
    \centering
    \includegraphics[scale=0.45]{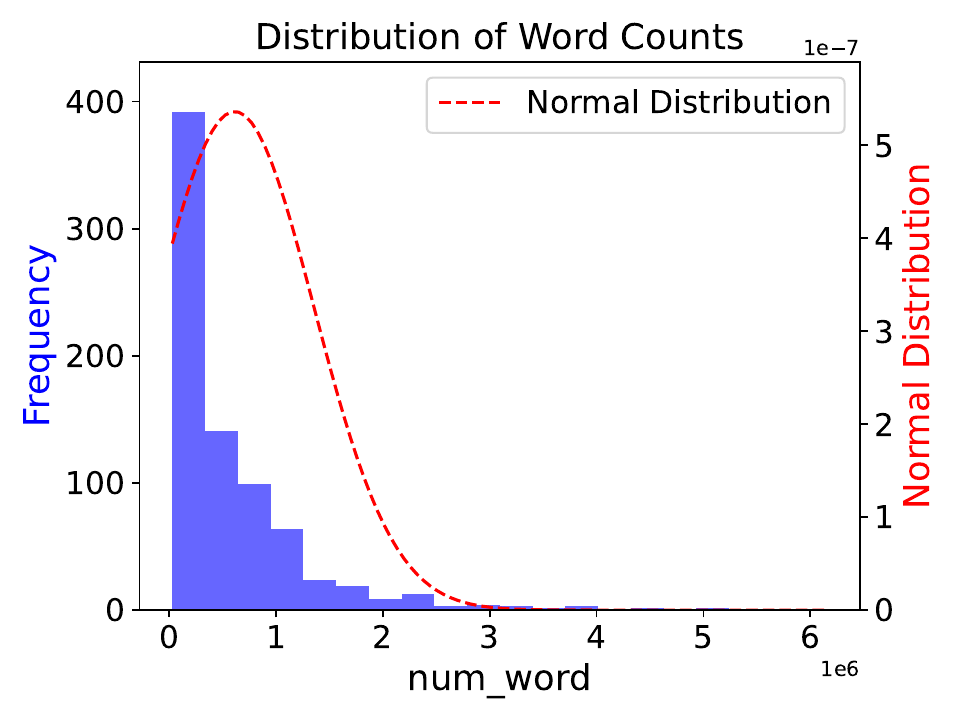}
    \vspace{-8pt}
    \caption{Distribution of total number of words in novels.
    The blue bars on the left represent the frequency distribution of word counts, while the red dashed line on the right represents the normal distribution curve of word counts.
    }
    \vspace{-12pt}
    \label{pic:num_word}
\end{figure}

\begin{figure}[!h]
    \centering
    \includegraphics[scale=0.45]{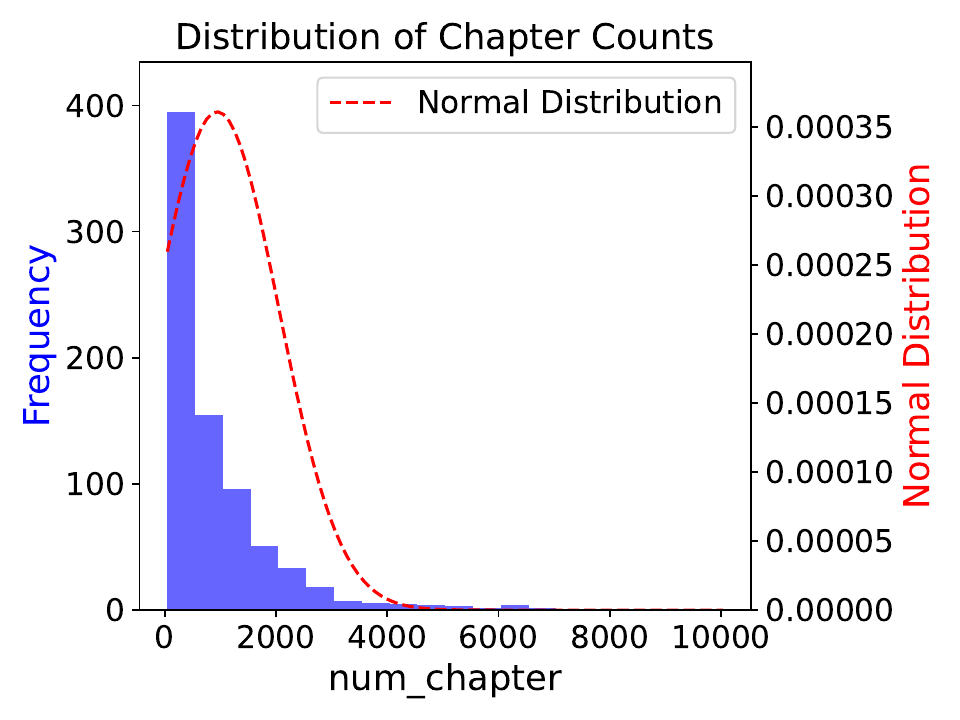}
    \vspace{-8pt}
    \caption{Distribution of total number of chapters in novels.
    The blue bars on the left represent the frequency distribution of chapter counts, while the red dashed line on the right represents the normal distribution curve of chapter counts.
    }
    \vspace{-12pt}
    \label{pic:num_chapter}
\end{figure}

We collected 800 web novels from the internet to form our dataset, covering various genres such as fantasy, suspense, romance, and modernity.
We analyzed the distribution of total word count and total number of chapters in these novels in the dataset, and plotted them as frequency distribution histograms and normal distribution plots.
Figure \ref{pic:num_word} and \ref{pic:num_chapter} respectively depict the statistical summaries of total word counts and total chapter counts.

\section{Details on Experiments}
\subsection{Model and API for baselines}
In this section, we present detailed descriptions of the models and experimental parameters used for each baseline in the experiments.

\subsubsection{Baselines Modification}
DOC, RE3, and Rolling-GPT are three baselines that take English novel premises as input and generate English novel texts.
These generated novels are then translated into Chinese using the DeepL deep learning translation package for both manual and automatic evaluations in comparison with Ex3.
Our intention behind this approach is to minimize any potential impact on the original performance of the baselines.
Both DOC and RE3 employ the GPT3-175B model for generating novel text while adhering to the original settings.

% The source code of DOC and RE3 employ several auxiliary models, such as the sentence encoder model, DPR model, and entailment model, etc., to assist in the generation of final novels.
% Since these models are designed only for English, therefore the original versions of DOC and RE3 do not support generating Chinese novels.
% Directly translating the English prompts of DOC and RE3 into Chinese does not enable the generation of Chinese novels.
% Replacing other models with their corresponding Chinese versions is also not feasible because we cannot guarantee that the performance of these two methods will not be affected after the replacement of models.
% To ensure fairness and minimize the impact on baseline performance, we have chosen not to modify the source code of DOC and RE3.
% Both DOC and RE3 employ the GPT3-175B model for generating novel text while adhering to the original settings.

During the novel generation process, we first input the English novel premise into the Plan module of DOC to generate a novel outline.
Then, the novel outline is respectively fed into the subsequent modules of DOC and RE3 to generate the main content of their respective novels.
However, RE3 requires an additional plan conversion module to convert the original outline into a format that RE3 can process.
We also utilize the Rolling-GPT code from DOC's GitHub repository as another baseline to reduce extraneous variables in the experiments.
Rolling-GPT also employs GPT3-175B and takes the outline generated by DOC to generate novels.

The R\scalebox{0.9}{ECURRENT}GPT method is a more convenient and flexible baseline as it utilizes only one LLM for generating novel text, primarily controlled through prompts in the source code.
We translated all the English prompts in the baseline source code to Chinese and utilized BaiChuan2-13B-Chat as the LLM for generating novel text, ensuring the performance of R\scalebox{0.9}{ECURRENT}GPT while supporting the generation of Chinese novels.
In addition to the aforementioned modifications, we have also added a preprocessing module at the input end of R\scalebox{0.9}{ECURRENT}GPT, which is to use the LLM to convert the input Chinese premise into a format acceptable by R\scalebox{0.9}{ECURRENT}GPT.

\subsubsection{Why Chooce Translation}
The source code of DOC and RE3 employ several auxiliary models, such as the sentence encoder model, DPR model, and entailment model, etc., to assist in the generation of final novels.
Since these models are designed only for English, therefore the original versions of DOC and RE3 do not support generating Chinese novels.
Directly translating the English prompts of DOC and RE3 into Chinese does not enable the generation of Chinese novels.
Replacing other models with their corresponding Chinese versions is also not feasible because we cannot guarantee that the performance of these two methods will not be affected after the replacement of models.
To ensure fairness and minimize the impact on baseline performance, we have chosen not to modify the source code of DOC and RE3.

The reason why we do not use RE3 or DOC data to train the novel writing model is that RE3 and DOC directly utilize pre-trained GPT3-175B to generate outlines and content of the novels without fine-tuning.
Their data is solely used for training auxiliary models such as consistency discriminators and relevance discriminators.

Considering that translation does not affect the plot and writing of the novels, we believe using the official implementation of them and adopting the translation approach for evaluation is the most proper method for comparison.
For R\scalebox{0.9}{ECURRENT}GPT, which does have Chinese language capabilities, we choose to directly generate Chinese novels for evaluation purposes.
We believe we have tried our best to make the comparison fair.

\subsection{Human Study Details}\label{appendix:human_study_details}
In the process of human evaluation, we maintain the same questionnaire setup as previous works and follow the pairwise approach in the experiments.
We distribute questionnaires to annotators for human evaluation.
Below we present the experimental details of our human study.

\subsubsection{Experiment Annotators Details}
We recruit 20 graduate students through an open channel within the university for human study.
The annotators who participate in the experiment all have a bachelor's degree or above.
All of them are native Chinese speakers and were not previously aware of this project, nor did they have any vested interests in it.

% Before receiving the questionnaire, each annotator is not aware of its contents and is required to answer it according to their actual feeling.
% The salary for each annotator is \$xx per hour, which includes the compensation for completing the survey.

During the experiment, annotators are asked to fill out questionnaires based on their genuine feelings.
% The annotators are required to fill out the questionnaire within specific time limits.
Follow the previous works,  we provide annotators with full text and allow them sufficient time to read the complete novels.
For the main experiments, the response time for each questionnaire in the medium-length novel experiment should be within 5 minutes, while for the long novel and ablation experiments, the response time should be within 10 minutes.

\subsubsection{Experiment Questionnaire Template}
\begin{figure*}[!t]
    \centering
    \begin{subfigure}{\linewidth}
        \centering
        \includegraphics[scale=0.26]{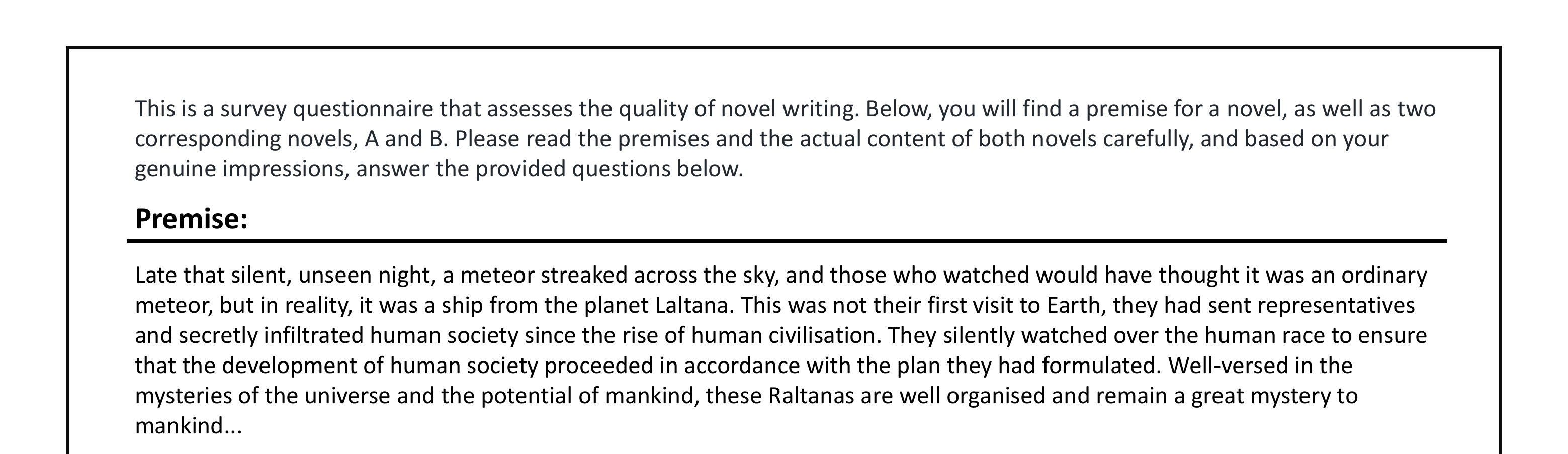}
        \phantomsubcaption
        \label{fig:sub1}
    \end{subfigure}
    \vspace{3pt}
    \begin{subfigure}{\linewidth}
        \centering
        \includegraphics[scale=0.26]{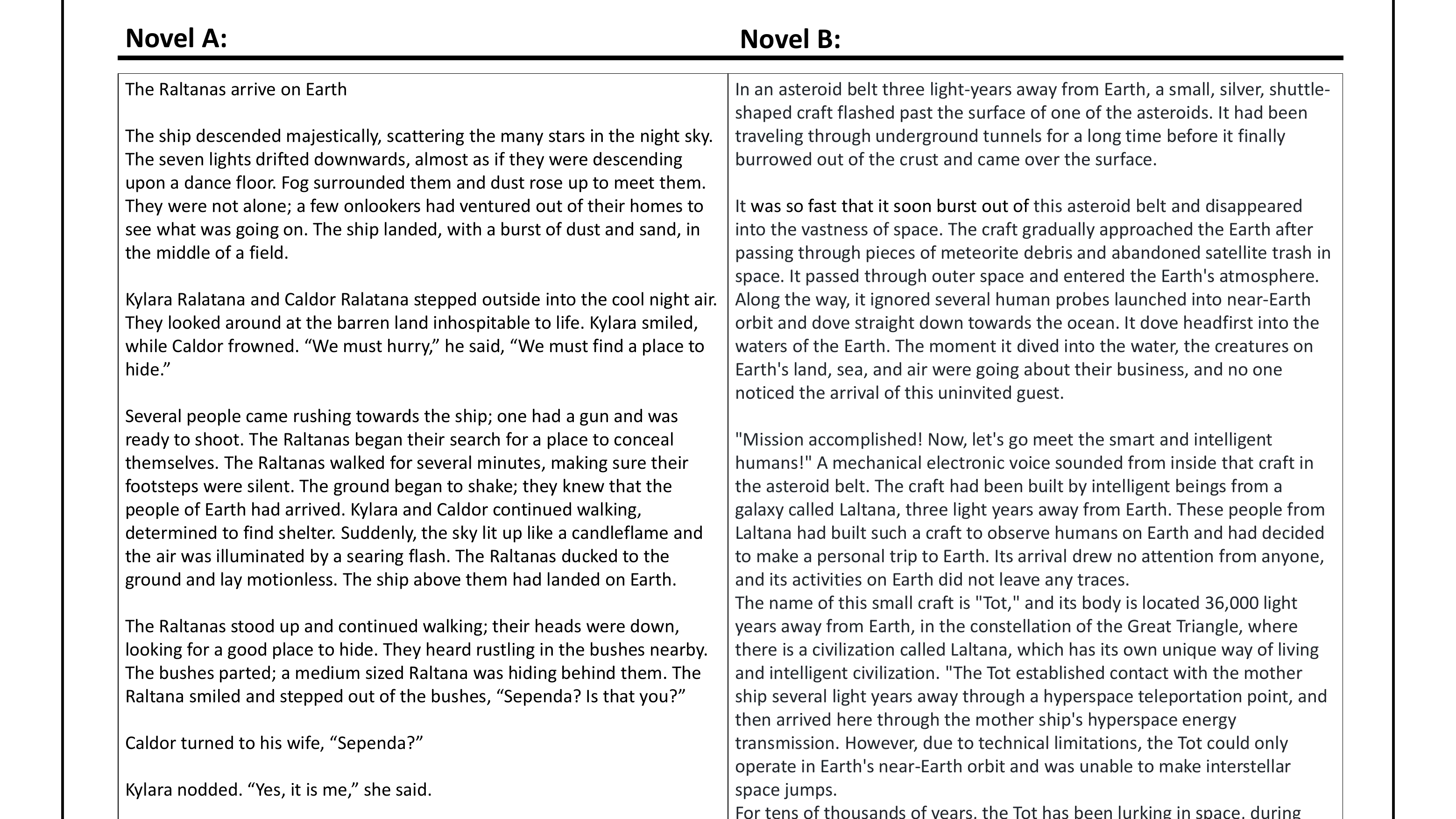}
        \phantomsubcaption
        \label{fig:sub2}
    \end{subfigure}
    \vspace{3pt}
    \begin{subfigure}{\linewidth}
        \centering
        \includegraphics[scale=0.26]{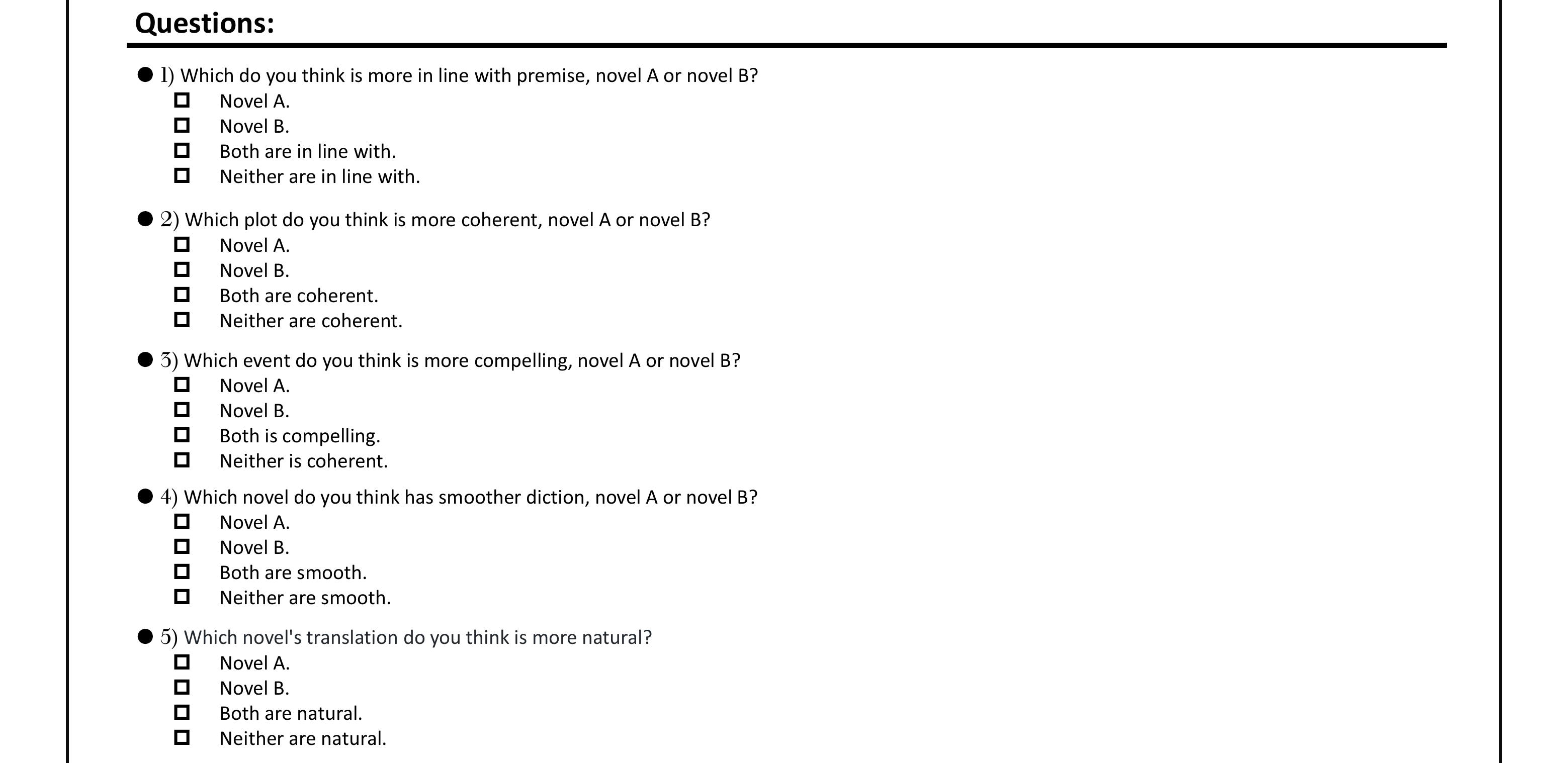}
        \phantomsubcaption
        \label{fig:sub3}
    \end{subfigure}
    \vspace{3pt}
    \begin{subfigure}{\linewidth}
        \centering
        \includegraphics[scale=0.26]{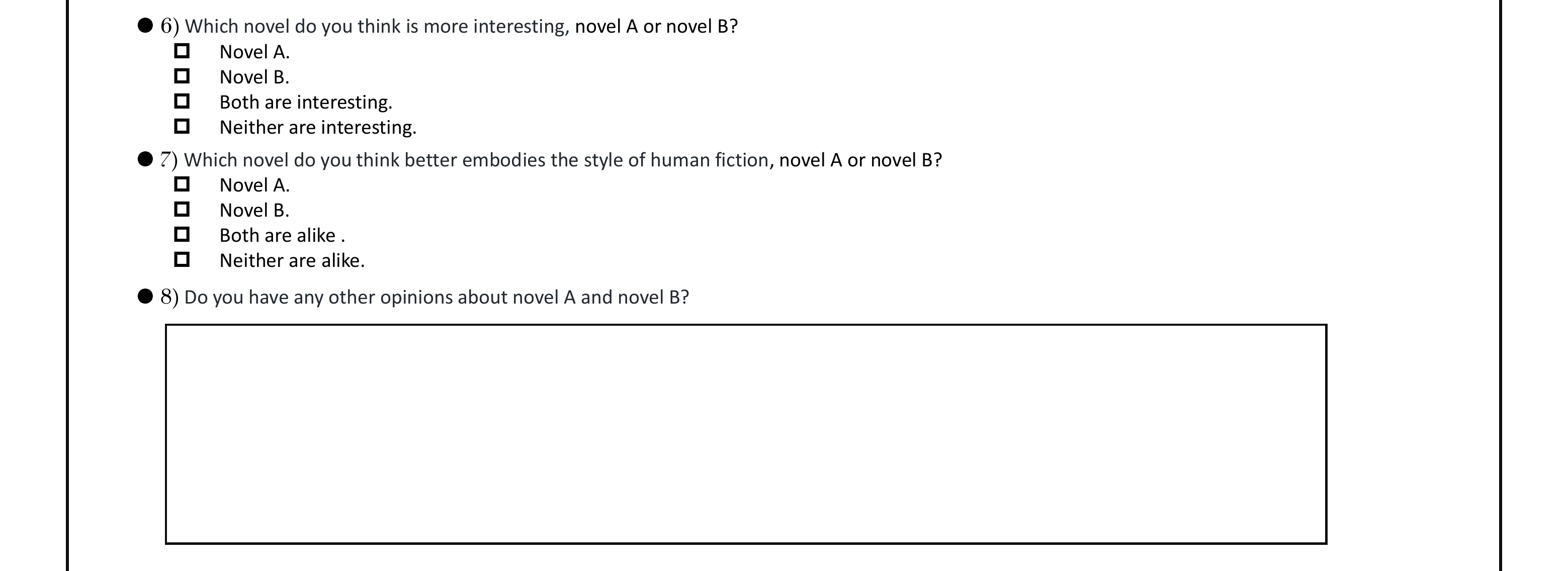}
        \phantomsubcaption
        \label{fig:sub3}
    \end{subfigure}
    
    \caption{An example of a questionnaire format used in our experiment. Due to the lengthy nature of the novel, only a portion is displayed.}
    \label{fig:questionnaire_template}
\end{figure*}

The questionnaire templates for the main experiments and ablation experiments are shown in Figure \ref{fig:questionnaire_template}.
Each questionnaire consists of instructions, a novel premise, two corresponding short stories A and B (where one is generated by Ex3 and the other by the baseline or ablation), as well as questions corresponding to the metrics for Human evaluation.

\subsubsection{Experiment With Human-authored Novels}
\label{vs_human}
Following the previous works, we only presented the results of the human evaluation for Ex3 and other automatic novel generation methods in the main text.
We expect that there is still a gap between existing auto-generated novel and novel written by human experts.

To ensure the completeness of the experiment, we supplemented the human evaluation experiments with a set of Ex3-generated novels and human-authored novels.
The Ex3 group novels were sourced from the medium-length experiments in the main text, while the human-authored group novels were based on the original texts corresponding to the premises used by the Ex3 group.
The other settings of the experiment remained consistent with those of the medium-length experiment.

As shown in Table \ref{tab:human_comparison_experiment.}, we find that Ex3 still has a gap with humans, which is consistent with out intuition, but it is close to human in Event Diction and Interesting metrics.

\subsubsection{Experiment Data Explanation}
We follow the pairwise approach used in previous works to compare our proposed method with the baselines, such as "Ex3 VS. DOC" and "Ex3 VS. RE3". 
The pairwise comparison means that annotators have to decide which novel is better on a certain metric or select neither or both entries if it can not be distinguished. 

The score is actually a percentage and is calculated as follows (number of annotators who believe the novel generated by method A meets the metric + number of annotators who believe both novels generated by different methods meet the metric) / total sample size * 100.
The two scores within a single cell of the table are co-dependent.
Obviously, this scores will change depending on the method to be compared, which is why the scores in our experimental results differ from previous articles. 
Therefore, it is meaningless to compare the numerical values between baselines in the human evaluation results across different papers.

We employed language different from previous papers in our comparisons, but we believe this does not have a significant impact on the method's capability.

\section{Annotator Agreement}
The experimental results of human evaluation rely on the judgments of annotators.
Although we have tried our best to ensure that annotators fill out the questionnaires truthfully and reasonably, the final judgments on different metrics are still quite subjective, and there may be poor agreement among annotators.
Following previous works, we compared annotator agreement by calculating Fleiss’ kappa under different experimental conditions.
The results can be seen in Tables \ref{tab:agreement_medium.}, \ref{tab:agreement_long.}, and \ref{tab:agreement_ablation.}, where overall agreement on different metrics is generally low, with some values even below 0.

\section{Analysis of Intermediate Results}
\label{inter_res}
Without proper quantitative metrics of whether these modules used in Ex3 work properly, it is difficult to evaluate them directly.
That may be the reason why previous works like RE3 and DOC don't directly evaluate the modules of their methods.
To comprehensively understand the effectiveness of each module in Ex3, we conducted a series of human studies to evaluate the intermediate steps.

\paragraph{Whether the partition in Figure \ref{GrouptheTextbySimilarity} makes sense}
We randomly sampled 10 text grouping examples and asked the annotators to judge whether the grouping was appropriate according to whether the text grouping completely ensured that the coherent plot was grouped into the same part.
We take the average of different annotators and different samples, and the appropriate proportion was \textbf{77.5\%}.

\paragraph{Whether the summary in Figure \ref{ChapterSummarizing} is faithful}
We randomly sampled 10 text-summary examples.
Each sample contains successive text groups and their summaries, as well as the overall summary.
Annotators are asked to judge whether the summary is sufficiently concise (conciseness) and whether the summary fully includes the main plot in the original text (completeness).
We take the average of different annotators and different samples, and the results are shown in Table \ref{tab:summary_faithfulness.}.

\begin{table}[H]
\centering
\begin{tabular}{c c}
\textbf{\makecell[c]{Conciseness}} & \textbf{\makecell[c]{Completeness}}\\ 
\hline
92.5\% & 80.0\% \\
\end{tabular}
\vspace{-8pt}
\caption{\label{tab:summary_faithfulness.}
The percentage of annotators who consider the summary to be sufficiently concise (conciseness) and whether the summary fully includes the main plot in the original text (completeness). 
It can be observed that the results of our summaries are widely accepted by most annotators.
}
\vspace{-14pt}
\end{table}

\paragraph{Whether the summaries in Figure \ref{RecursivelySummarizing} are of different levels of details}
We sampled 10 sets of summaries, each containing 4-6 different levels of summaries, and asked annotators to judge whether these summaries are of different levels of detail.
We took the average of different annotators and different samples.
In the end, \textbf{92\%} said there was a difference.

We also introduce a harder task that asks annotators to rank these summaries according to their degree of detail.
We calculate the Kendall tau rank correlation coefficient between the sorting given by the annotators and the real sorting.
The value range of this coefficient is [-1,1].
The larger the coefficient is, the stronger the correlation degree of the sorting is.
We take the average of different annotators and different samples, the average Kendall tau rank correlation coefficient is 0.1777.
Thus, the summaries in Figure \ref{RecursivelySummarizing} are of different levels of detail.

%% Ex3 and human-written
\begin{table*}
\centering
\begin{tabular}{l c c c c c c c}
\hline
\textbf{\makecell{Method}} & \textbf{Relevant} & \textbf{Coherent} & \textbf{Event} & \textbf{Diction} & \textbf{Transition} & \textbf{Interesting} & \textbf{Human-like}\\ 
\hline
Ex3 & 50.00 & 36.67 & 40.00 & 50.00 & 16.67 & 53.33 & 56.67 \\
Human & \textbf{90.00} & \textbf{96.67} & 52.50 & 60.00 & \textbf{96.67} & 66.67 & \textbf{90.00} \\
\hline
\end{tabular}
\vspace{-8pt}
\caption{\label{tab:human_comparison_experiment.}
Pair-wise comparison of Ex3 with Human-authored novels, respectively.
The horizontal axis of represents different methods being compared, the vertical axis represents the metrics being compared, and the table values indicate the percentage of annotators who believe the method's novels align with the metrics.
Results in different comparisons are not comparable with each other. 
Bold indicates significance with $p<0.05$. 
}
\vspace{-14pt}
\end{table*}

%% medium-length agreement
\begin{table*}
\centering
\renewcommand{\arraystretch}{2} % 设置行高的倍数
\begin{tabular}{c c c c c c c c}
\hline
\textbf{\makecell{Method}} & \textbf{Relevant} & \textbf{Coherent} & \textbf{Event} & \textbf{Diction} & \textbf{Transition} & \textbf{Interesting} & \textbf{Human-like}\\ 
\hline
Ex3  vs. DOC & 0.1554 & 0.0489 & 0.1992 & -0.1195 & 0.1331 & 0.0122 & 0.1331 \\
\hline
Ex3  vs. RE3 & 0.1992 & -0.2716 & -0.1755 & -0.1777 & -0.0920 & -0.1852 & -0.1279 \\
\hline
\makecell[c]{Ex3  vs.\\ R\scalebox{0.9}{ECURRENT}GPT} & -0.0824 & -0.2692 & -0.1263 & -0.2849 & 0.0443 & 0.0969 & -0.0012 \\
\hline
\makecell[c]{Ex3  vs.\\ Rolling-GPT} & -0.2615 & -0.1784 & -0.0513 & -0.0513 & 0.1330 & -0.1124 & -0.1124 \\
\hline
\end{tabular}
\vspace{-8pt}
\caption{\label{tab:agreement_medium.}
Fleiss’ kappa for agreement on individual metric annotations in pairwise comparisons between Ex3
and different baselines under the scale of medium-length novels.
Overall the agreement is relatively poor.
}
\vspace{-14pt}
\end{table*}

%% long-length agreement
\begin{table*}
\centering
\renewcommand{\arraystretch}{2} % 设置行高的倍数
\begin{tabular}{c c c c c c c c}
\hline
\textbf{\makecell{Method}} & \textbf{Relevant} & \textbf{Coherent} & \textbf{Event} & \textbf{Diction} & \textbf{Transition} & \textbf{Interesting} & \textbf{Human-like}\\ 
\hline
Ex3  vs. DOC & 0.4253 & -0.2201 & -0.1475 & -0.1655 & -0.1952 & -0.1458 & -0.1384 \\
\hline
Ex3  vs. RE3 & 0.0000 & -0.0751 & -0.0417 & -0.1576 & -0.2295 & 0.0586 & -0.0563 \\
\hline
\makecell[c]{Ex3  vs.\\ R\scalebox{0.9}{ECURRENT}GPT} & -0.1925 & 0.0102 & -0.3014 & -0.1538 & -0.3235 & -0.2000 & 1.0000 \\
\hline
\makecell[c]{Ex3  vs.\\ Rolling-GPT} & 0.0230 & -0.0526 & -0.2054 & -0.0145 & -0.1111 & -0.0372 & -0.1215 \\
\hline
\end{tabular}
\vspace{-8pt}
\caption{\label{tab:agreement_long.}
Fleiss’ kappa for agreement on individual metric annotations in pairwise comparisons between Ex3
and different baselines under the scale of long-length novels.
Overall the agreement is relatively poor.
}
\vspace{-14pt}
\end{table*}

%% ablation agreement
\begin{table*}
\centering
\renewcommand{\arraystretch}{2} % 设置行高的倍数
\begin{tabular}{c c c c c c c c}
\hline
\textbf{\makecell{Method}} & \textbf{Relevant} & \textbf{Coherent} & \textbf{Event} & \textbf{Diction} & \textbf{Transition} & \textbf{Interesting} & \textbf{Human-like}\\ 
\hline
\makecell[c]{Ex3  vs.\\ NO-TRAIN} & -0.2844 & 0.1241 & -0.1570 & -0.0309 & -0.4400 & -0.2844 & 0.0361 \\
\hline
\makecell[c]{Ex3  vs.\\ NO-ENTITY} & 0.3966 & -0.0601 & -0.0889 & -0.0645 & 0.0106 & -0.3089 & 0.0238 \\
\hline
\end{tabular}
\vspace{-8pt}
\caption{\label{tab:agreement_ablation.}
Fleiss’ kappa for agreement on individual metric annotations in pairwise comparisons between Ex3
and other ablation experiments.
Overall the agreement is relatively poor.
}
\vspace{-14pt}
\end{table*}

% Questionnaire Template
% \begin{figure*}[!t]
%     \centering
%     \includegraphics[scale=0.26]{Pics/Questionnaire1.pdf}
% \end{figure*}

% \begin{figure*}[!t]
%     \centering
%     \includegraphics[scale=0.26]{Pics/Questionnaire2.pdf}
% \end{figure*}

% \begin{figure*}[!t]
%     \centering
%     \includegraphics[scale=0.26]{Pics/Questionnaire3.pdf}
% \end{figure*}

% \begin{figure*}[!t]
%     \centering
%     \includegraphics[scale=0.26]{Pics/Questionnaire4.pdf}
% \end{figure*}

% \begin{figure*}[!t]
%     \centering
%     \includegraphics[scale=0.26]{Pics/Questionnaire5.pdf}
%     \caption{An example of a questionnaire format used in our experiment. Due to the lengthy nature of the novel, only a portion is displayed.}
%     \label{fig:questionnaire_template}
% \end{figure*}

\end{document}